\pdfoutput=1
 
\documentclass[11pt]{article}

\usepackage[final]{acl}

\usepackage{times}
\usepackage{latexsym}
\usepackage{caption}
\usepackage{subfigure}
\usepackage{physics}
\usepackage{amsmath}
\usepackage{tikz}
\usepackage{yhmath}
\usepackage{cancel}
\usepackage{color}
\usepackage{siunitx}
\usepackage{array}
\usepackage{multirow}
\usepackage{amssymb}
\usepackage{gensymb}
\usepackage{tabularx}
\usepackage{extarrows}
\usepackage{booktabs}
\usepackage{tabularx}
\usepackage{pgfplots}
\usetikzlibrary{fadings}
\usetikzlibrary{patterns}
\usetikzlibrary{shadows.blur}
\usetikzlibrary{shapes}
\usepackage{colortbl}
\usepackage{xcolor}
\usepackage{tipa}
\definecolor{lgreen}{rgb}{0.937,0.992,0.929}
\definecolor{dgreen}{rgb}{0.470,0.650,0.369}
\definecolor{lblue}{rgb}{0.902,0.933,1.000}
\definecolor{llblue}{RGB}{235,242,247}
\definecolor{dblue}{RGB}{165,195,229}
\definecolor{lyellow}{RGB}{252,234,185}
\definecolor{dodgerblue}{rgb}{0.117,0.564,1.000}
\definecolor{orangered}{rgb}{1.000,0.270,0.000}
\definecolor{nasdaqup}{rgb}{0.000,0.654,0.356}
\usepgfplotslibrary{groupplots} 
\usepgfplotslibrary[groupplots]
\usetikzlibrary{pgfplots.groupplots}
\usetikzlibrary[pgfplots.groupplots]
\usetikzlibrary{fit}
\usetikzlibrary{backgrounds}
\usepackage[T1]{fontenc}

\usepackage[utf8]{inputenc}

\usepackage{microtype}

\usepackage{inconsolata}

\usepackage{soul}
\title{
Leveraging Unit Language Guidance to Advance Speech Modeling\\ in Textless Speech-to-Speech Translation
}

\author{Yuhao Zhang$^{1,2}$, Xiangnan Ma$^1$, Kaiqi Kou$^1$, Peizhuo Liu$^1$, Weiqiao Shan$^1$,\\ \bf{Benyou Wang$^{2}$, Tong Xiao$^{1,4}$\thanks{\ \ Corresponding author.}, Yuxin Huang$^{3}$, Zhengtao Yu$^{3}$ and Jingbo Zhu$^{1,4}$}\\
    $^{1}$ Northeastern University, Shenyang, China\\
    $^{2}$ The Chinese University of Hong Kong, Shenzhen, China \\
    $^{3}$ Kunming University of Science and Technology, Kunming, China \\
    $^{4}$NiuTrans Research, Shenyang, China \\
    {\tt
    yoohao.zhang@gmail.com, \{xiaotong, zhujingbo\}@mail.neu.edu.cn
    }
}

\begin{document}
\maketitle
\begin{abstract}

The success of building textless speech-to-speech translation (S2ST) models has attracted much attention. However, S2ST still faces two main challenges: 1) extracting linguistic features for various speech signals, called cross-modal (CM), and 2) learning alignment of difference languages in long sequences, called cross-lingual (CL). We propose the unit language to overcome the two modeling challenges. The unit language can be considered a text-like representation format, constructed using $n$-gram language modeling. We implement multi-task learning to utilize the unit language in guiding the speech modeling process. Our initial results reveal a conflict when applying source and target unit languages simultaneously. We propose task prompt modeling to mitigate this conflict. We conduct experiments on four languages of the Voxpupil dataset. Our method demonstrates significant improvements over a strong baseline and achieves performance comparable to models trained with text.

\end{abstract}

\section{Introduction}

The Speech-to-Speech Translation (S2ST) task aims to generate target speech according to the source speech, which can significantly improve communication efficiency between speakers of different languages. Conventional methods apply the cascade method that uses automatic speech recognition (ASR), machine translation (MT), and text-to-speech (TTS) models \cite{vidal1997finite, casacuberta2004some, aguero2006prosody}. This strategy always suffers from error propagation and high latency, thus researchers turn to investigate direct S2ST \cite{jia19_interspeech, lee-etal-2022-direct, inaguma-etal-2023-unity}. The direct S2ST models the source speech and generates the unit or spectrum of the target speech within one model. Inspired by the direct S2ST paradigm, \citet{lee-etal-2022-textless} proposed \textit{textless} S2ST, namely achieving speech-to-speech transformation with the constraint of unit or VQ-VAE token. This method eliminates the need for any labeled text and is very valuable for languages without text-writing systems.

\input{ul_figure}

However, there are still two challenging problems for textless S2ST \cite{lee-etal-2022-direct, Jia2021Translatotron2H}: 1) how to learn the acoustic and linguistic features from the varying and continuous audio signal without transcription (calling it cross-modal modeling, CM), and 2) how to achieve alignment between languages along the long sequence without translated text pairs (calling it cross-lingual modeling, CL). For the former problem, previous works propose the quantization method to improve the discrete unit or token, aiming to be more suitable for cross-modal and cross-lingual modeling \cite{zhang2020uwspeech, lee-etal-2022-direct, 10096797}. In the latter case, some masked language models or denoising auto-encoder methods were used to improve cross-lingual modeling \cite{popuri22_interspeech, chen-etal-2023-speech}. Though these methods show improvement in textless S2ST, the CL and CM challenges have not been adequately addressed due to the lack of guidance from text.

We design a format called the unit language, serving as a transcription of speech to enhance CM and CL modeling.
The unit language consists of unit words which are merged from several contiguous speech units. Each unit word is generated based on $n$-gram language modeling. This strategy searches for the maximum probability of unit words for every unit sequence and does not rely on any labeled data. This novel representation can serve as an alternative to text during the modeling process, addressing the challenges posed by textless training. As shown in Figure \ref{Unit_language}, the unit language is implicitly aligned with the real text. We further implement multi-task learning based on the source and target unit language to guide CM and CL modeling respectively.

We conducted experiments on the Voxpupil dataset \cite{wang-etal-2021-voxpopuli}. Our method achieved an average improvement of 1.2 BLEU over the strong baseline. Furthermore, it demonstrates performance comparable to the S2ST model trained with text. This shows that the unit language can serve as an effective alternative to text in the speech modeling process. Our further analysis reveals that CM and CL processing based on unit language have distinct impacts, but both are essential for S2ST. CM processing filters noise in speech, while CL processing helps capture semantic information for translation. However, the negative impact of CL on the effectiveness of CM prevents the simultaneous application of both methods from achieving consistent improvements. To address this issue, we proposed task prompt modeling to mitigate the conflict. Final results demonstrate that our model achieves new state-of-the-art performance on the textless S2ST task using the Voxpupil dataset\footnote{Code: \href{https://github.com/xiaozhang521/Unit_Language}{https://github.com/xiaozhang521/Unit\_Language}}.

\section{Method}

The philosophy of this paper is to utilize a text-like format, namely the unit language, to enhance speech modeling. To use the unit language, we introduce two additional decoders and employ multi-task learning to guide the modeling process. We propose task prompt modeling as a mitigation strategy to address task conflicts in multi-task learning.

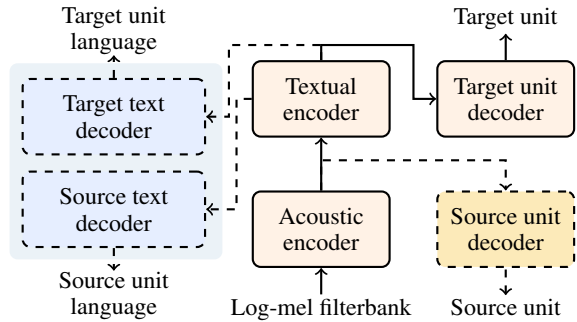
\begin{figure}[t]
    \centering
    \small
      \begin{tikzpicture} [scale=1.0]
        \node(ae) at (0,0) [rectangle, draw=black, fill=orange!10, rounded corners=3pt, thick, minimum width=1.8cm,minimum height=1cm,align=center] {Acoustic\\encoder};
        \node(fb) at ([yshift=-0.3cm]ae.south) [rectangle, align=center,anchor=north] {Log-mel filterbank};
        \node(te) at ([yshift=0.7cm]ae.north) [rectangle, draw=black, fill=orange!10, rounded corners=3pt, thick, minimum width=1.8cm,minimum height=1cm,align=center,anchor=south] {Textual\\encoder};
        \node(tud) at ([xshift=0.6cm]te.east) [rectangle, draw=black, fill=orange!10, rounded corners=3pt, thick, minimum width=1.8cm,minimum height=1cm,align=center,anchor=west] {Target unit\\decoder};
        \node(tu) at ([yshift=0.3cm]tud.north) [rectangle, align=center,anchor=south] {Target unit};
        \node(sud) at ([xshift=0.6cm]ae.east) [rectangle, dashed, draw=black, fill=lyellow, rounded corners=3pt, thick, minimum width=1.8cm,minimum height=1cm,align=center,anchor=west] {Source unit\\decoder};
        \node(su) at ([yshift=-0.3cm]sud.south) [rectangle, align=center,anchor=north] {Source unit};
        
        \node(backgound) at (-2.7,0.9) [draw=none, fill=llblue, rounded corners=3pt, minimum width=2.8cm,minimum height=2.6cm, align=center] {};
        \node(sd) at ([xshift=-0.6cm, yshift=-0.6cm]$(ae.west)!0.5!(te.west)$) [rectangle, draw=black, fill=lblue, rounded corners=3pt, thick, minimum width=2.4cm,minimum height=1cm,align=center,anchor=east,dashed] {Source text\\decoder};
        \node(td) at ([yshift=0.2cm]sd.north) [rectangle, draw=black, fill=lblue, rounded corners=3pt, thick, minimum width=2.4cm,minimum height=1cm,align=center,anchor=south,dashed] {Target text\\decoder};
        \node(sul) at ([yshift=-0.2cm]sd.south) [rectangle, align=center,anchor=north] {Source unit\\language};
        \node(tul) at ([yshift=0.2cm]td.north) [rectangle, align=center,anchor=south] {Target unit\\language};
        
        \draw[->,thick]([yshift=-0.05cm]fb.north)--(ae.south);
        \draw[->,thick](tud.north)--([yshift=0.05cm]tu.south);
        \draw[->,thick,dashed]([yshift=-0.02cm]sud.south)--([yshift=-0.06cm]su.north);
        \draw[->,thick](ae.north)--(te.south);
        \draw[->,thick](te.north)--([yshift=0.2cm]te.north)--([yshift=0.2cm,xshift=-0.3cm]tud.north -| tud.west)--([xshift=-0.3cm]tud.west)--(tud.west); 
        \draw[->,thick,dashed](te.west)--([xshift=0.4cm]te.west-|sd.east)--([xshift=-0.2cm]te.west|-sd.east)--(sd.east);
        \draw[->,thick,dashed](ae.north)--([yshift=0.4cm]ae.north)--([yshift=0.4cm]sud.north)--(sud.north);
        \draw[->,thick,dashed](te.north)--([yshift=0.2cm]te.north)to([yshift=0.2cm,xshift=0.3cm]te.north -| td.east)--([xshift=0.3cm]td.east)--(td.east);
        
        \draw[->,thick,dashed](sd.south)--([yshift=-0.1cm]sul.north);
        \draw[->,thick,dashed](td.north)--([yshift=0.1cm]tul.south);
        
      \end{tikzpicture}
      \caption{Architecture of textless S2ST. The dashed line part will be removed during inference. The filterbank features serves as the only input to the network. Blue modules are used to guide the modeling process. }
      \label{architecture}
  \end{figure}

  
\subsection{Unit Language Construction}

We propose a method to construct the unit language based on language modeling processing. Formally, given a unit sequence \{$u_1, u_2, ..., u_n$\}, our goal is to convert it into a word sequence \{$w_1, w_2, ..., w_m$\}, where each $w$ consists of at most $K$ continuous units and $K$ is a hyper-parameter. Considering an arbitrary sub-sequence $u_{\left[1:i\right]}=\{u_1, u_2, ..., u_i\}$, we can generate a corresponding sequence $w_{\left[1:j\right]}=\{w_1, w_2, ..., w_j\}$. Considering the original unit can be various, we apply the \textbf{norm unit} \cite{lee-etal-2022-textless}, which is produced by a model trained with data from a specific speaker. The norm unit is less noisy and easier to learn. According to the process of language modeling, the probability of $w_{\left[1:j\right]}$ can be calculated by:
\begin{equation}
    P(w_{\left[1:j\right]})=P(w_1)P(w_2|w_1)...P(w_j|w_1...w_{j-1}).
\end{equation}
When the $P(w_{\left[1:j\right]})$ reaches its maximum, which is the maximum likelihood, we can consider $w_{\left[1:j\right]}$ as the optimal unit language sequence to represent $u_{\left[1:i\right]}$. We use $\pi(u_{\left[1:i\right]})$ to denote the optimal conversion path as follows:
\begin{equation}
\pi(u_{\left[1:i\right]})=w^*_{\left[1:j\right]} = \mathop{\arg\max}\limits_{w_{\left[1:j\right]}}(P(w_{\left[1:j\right]})).
\end{equation}
\paragraph{1-gram} Our final goal is to find $\pi$ for any unit sequence. To make the deduction process clear, we first consider the circumstance of conditional independence, namely 1-gram. Then we can get:
\begin{equation}
\begin{aligned}
P(w_{\left[1:j\right]})&\approx P(w_1)P(w_2)...P(w_j)\\
&=\mathrm{exp}(\sum_{t}^{j}\mathrm{log}P(w_t)).
\end{aligned}
\end{equation}
Considering that $w_j$ consists of a maximum of $K$ continuous units, we can derive the following recursion formula for $w_j$ and $w_{\left[1:j-1\right]}$:
\begin{equation}
\begin{aligned}
    &\mathop{\max}(\sum_{t}^{j}\mathrm{log}P(w_t))=\\
    &\mathop{\max}\left\{ 
    \begin{aligned}
        &\mathrm{log}P(\pi(u_{\left[1:i-1\right]}))+\mathrm{log}P(u_i),\\
        &\mathrm{log}P(\pi(u_{\left[1:i-2\right]}))+\mathrm{log}P(u_{\left[i-1,i\right]}),\\
        &...\\
        &\mathrm{log}P(\pi(u_{\left[1:i-K\right]}))+\mathrm{log}P(u_{\left[i-K+1,i\right]})
    \end{aligned}
    \right\}.
\end{aligned}
\end{equation}
For any $k \leq K$, we can use the frequency from the unit corpus to estimate $P(u_{\left[i-k+1:i\right]})$. Then, we get $k_i^*$ as the maximum element:
\begin{equation}
\small
k_i^*= \mathop{\arg\max}\limits_{k}(\mathrm{log}P(\underbrace{\pi(u_{\left[1:i-k\right]})}_{w^*_{\left[1:j-1\right]}}) 
+\mathrm{log}P(\underbrace{u_{\left[i-k+1,i\right]}}_{w_j}) ).
\end{equation}
For position $i$, we can determine $w_j$ using $k_i^*$ units, namely $w_j=u_{\left[i-k^*_i+1,i\right]}$. Then $w_{j-1}$ can be determined by $k_{i-k^*_i}^*$. After applying dynamic programming, the entire sequence $w^*_{\left[1:m\right]}$ can be obtained for any unit sequence $u_{\left[1:n\right]}$, and we consider the generated sequence $w^*_{\left[1:m\right]}$ as the unit language.

\paragraph{2-gram} For most languages, considering in-context information is necessary, and we can easily adapt our method to $n$-gram modeling. Considering the computational cost (refer to Appendix), we primarily use 2-gram modeling in this paper. Thus, equation (2) can be updated as follows:
\begin{equation}
\begin{aligned}
    &\pi(u_{\left[1:i\right]})
    = \mathop{\arg\max}\limits_{w_{\left[1:j\right]}}(P(w_{\left[1:j\right]}))\\
    &=\mathop{\arg\max}\limits_{w_{\left[1:j\right]}}(\sum_{t>1}^{j}\mathrm{log}P(w_t|w_{t-1})+\mathrm{log}P(w_{1}))
\end{aligned}
\end{equation}
Similarly, we can update the formula (4) as follows:
\begin{equation}
\begin{aligned}
    &\mathrm{max}(\sum_{t>1}^{j}\mathrm{log}P(w_t|w_{t-1}))=\\
    &\mathrm{max}\left\{
    \begin{aligned}
        &\mathrm{log}P(\pi(u_{\left[1:i-1\right]}))\\
        &+\mathrm{log}P(u_i|u_{\left[i-k^*_{i-1}:i-1\right]}),\\
        &\mathrm{log}P(\pi(u_{\left[1:i-2\right]}))\\
        &+\mathrm{log}P(u_{\left[i-1:i\right]}|u_{\left[i-k^*_{i-2}-1:i-2\right]}),\\
        &...\\
        &\mathrm{log}P(\pi(u_{\left[1:i-K\right]}))\\
        &+\mathrm{log}P(u_{\left[i-K+1:i\right]}|u_{\left[i-k^*_{i-K}-K+1:i-K\right]}) 
    \end{aligned}
   \right\} 
\end{aligned}
\end{equation}
For any $k \leq K$ and $l=k^*_{i-k}-k$, we still use the frequency to estimate the conditional probability as follows:
\begin{equation}
P(u_{\left[i-k+1:i\right]}|u_{\left[i-l+1:i-k\right]})=
\frac{P(u_{\left[i-l+1:i\right]})}{P(u_{\left[i-l+1:i-k\right]})}.
\end{equation}
\subsection{Architecture}
We improve the architecture of advanced textless S2ST model \cite{lee-etal-2022-textless} as shown in Figure \ref{architecture}. It converts source audio $a_{s}$ and target audio $a_{t}$ to discrete units $u_{s}$ and $u_{t}$, respectively. The source units $u_{s}$ and target units $u_{t}$ are produced by the pre-trained multilingual Hubert model \cite{hsu2021hubert}, and $u_{s}$ and $u_{t}$ serve as pseudo source and target text. We apply the norm unit \cite{lee-etal-2022-textless} as the training target to achieve strong baselines. 

To clarify the architecture, we define the baseline as consisting of four parts: acoustic encoder (A-Enc), textual encoder (T-Enc), source unit decoder (SU-Dec), and target unit decoder (TU-Dec). We add two additional decoders to process source and target unit languages. Since both decoders generate the text-like unit language, we call them the source text decoder (S-Dec) and target text decoder (T-Dec), respectively. Previous analysis shows that the representation is converted from speech to text in the textual encoder \cite{zhang-etal-2023-rethinking}. We choose the intermediate state of T-Enc as the input for S-Dec and T-Dec. 

At the inference stage, the audio features are fed into the A-Enc. After processing by A-Enc and T-Enc, the TU-Dec directly generates target units, which are subsequently passed to a vocoder to synthesize the target speech.

\begin{table}[t]
    \centering
    \small
    \setlength{\tabcolsep}{1.0mm}{
    \begin{tabular}{lclcc}
    \toprule
    Loss&Output&Training modules& \\
    \midrule
    $\mathcal{L}_{\mathrm{SU}}$&Source unit&\{A-Enc, SU-Dec\}& \\
    $\mathcal{L}_{\mathrm{TU}}$&Target unit&\{A-Enc, T-Enc, TU-Dec\}&\\
    $\mathcal{L}_{\mathrm{CM}}$&Source unit language&\{A-Enc, T-En$\mathrm{c}^{*}$, S-Dec\}&&\\
    $\mathcal{L}_{\mathrm{CL}}$&Target unit language&\{A-Enc, T-Enc, T-Dec\}&&\\
    $\mathcal{L}_{\mathrm{CM}^{'}}$&Source text&\{A-Enc, T-En$\mathrm{c}^{*}$, S-Dec\}&&\\
    $\mathcal{L}_{\mathrm{CL}^{'}}$&Target text&\{A-Enc, T-Enc, T-Dec\}&&\\
    \bottomrule
    \end{tabular}}
    \caption{Overview of all the tasks and modules. The input is filterbank for all the tasks. * denotes that the loss updates parts of parameters in the module.}
    \label{MTL}
\end{table}

\begin{table*}[t]
    \centering
    \small
    \setlength{\tabcolsep}{2.0mm}{
    \begin{tabular}{llllll}
    \toprule
     Models &\multicolumn{1}{c}{Es-En}&\multicolumn{1}{c}{Fr-En}&\multicolumn{1}{c}{En-Es}&\multicolumn{1}{c}{En-Fr}&\multicolumn{1}{c}{Avg.} \\
    \midrule
    Seamless \cite{barrault2023seamlessm4t} &34.7&35.9 &29.2&25.0&31.2\\
    Textless S2UT \cite{lee-etal-2022-textless} &18.9&19.9&22.7&18.7&20.1 \\
    S2UT$^*$ \cite{lee-etal-2022-textless} &19.4&19.7&21.8&18.9&20.0\\
    \midrule
    \multicolumn{6}{c}{\textit{w/}  Recognized text} \\
    \midrule
    Baseline (+ $\mathcal{L}_{\mathrm{TU}}$\&$\mathcal{L}_{\mathrm{SU}}$) & 19.1 & 20.3 & 23.0 & 18.8&20.3 \\
    \ \ + $\mathcal{L}_{\mathrm{CM}^{'}}$&19.7 ($+$0.6)&21.0 ($+$0.7)&23.9 ($+$0.9)& \textbf{20.8} ($+$2.0)&21.4 ($+$1.1)\\ 
    \ \ + $\mathcal{L}_{\mathrm{CL}^{'}}$&\textbf{20.3} ($+$1.2)&\textbf{21.2} ($+$0.9)&23.9 ($+$0.9)& 20.6 ($+$1.8)&\textbf{21.5} ($+$1.2)\\ 
    \ \ + $\mathcal{L}_{\mathrm{CM}^{'}}$\&$\mathcal{L}_{\mathrm{CL}^{'}}$&19.8 ($+$0.7)&20.8 ($+$0.5)&\textbf{24.0} ($+$1.0)&\textbf{20.8} ($+$2.0)& 21.4 ($+$1.1)\\ 
    \midrule
    \multicolumn{6}{c}{\textit{w/} Unit language} \\
    \midrule
    Baseline (+ $\mathcal{L}_{\mathrm{TU}}$\&$\mathcal{L}_{\mathrm{SU}}$) & 19.1 & 20.3 & 23.0 & 18.8&20.3 \\
    \ \ + $\mathcal{L}_{\mathrm{CM}}$& 19.2 ($+$0.1) & 20.7 ($+$0.4)&23.5 ($+$0.5)&19.5 ($+$0.7)&20.7 ($+$0.4)\\ 
    \ \ + $\mathcal{L}_{\mathrm{CL}}$&19.4 ($+$0.3)&21.0 ($+$0.7)&23.9 ($+$0.9)&19.9 ($+$1.1)&21.1 ($+$0.8)\\
    \ \ + $\mathcal{L}_{\mathrm{CM}}$\&$\mathcal{L}_{\mathrm{CL}}$&19.7 ($+$0.6)&21.0 ($+$0.7)&23.8 ($+$0.8)&20.4 ($+$1.6)&21.2 ($+$0.9)\\
    \ \ \ \ +Task prompt&\textbf{19.9} ($+$0.8)&\textbf{21.1} ($+$0.8)& \textbf{24.4} ($+$1.4)&\textbf{20.6} ($+$1.8)&\textbf{21.5} ($+$1.2)\\ 
    \bottomrule
    \end{tabular}}
    \caption{Performance on different datasets. Our baseline is a reproduction \cite{lee-etal-2022-textless}. * denotes that the model does not use the norm unit.}
    \label{main}
\end{table*}

\subsection{Multi-task Learning}

Training the textless S2ST model is challenging, and mainstream methods rely on multi-task learning. Specifically, the input is the filterbank feature of $a_{s}$, and the output of the acoustic encoder is used to predict the source unit $u_{s}$ as the training loss:
\begin{equation}
    \mathcal{L}_{\mathrm{SU}} = - \mathrm{log} P(u_{s}|a_{s}, \theta_{\mathrm{A\text{-}Enc}}, \theta_{\mathrm{SU\text{-}Dec}}).
\end{equation}
The output of the whole encoder (A-Enc and T-Enc) is used to predict the target unit by TU-Dec. The cross-lingual loss is denoted as:
\begin{equation}
    \mathcal{L}_{\mathrm{TU}} = - \mathrm{log} P(u_{t}|a_{s}, \theta_{\mathrm{A\text{-}Enc}}, \theta_{\mathrm{T\text{-}Enc}}, \theta_{\mathrm{TU\text{-}Dec}}).
\end{equation}

To improve the effect of CM, we use the unit language as the transcription of each source audio, denoted $\widetilde{u_s}$. The S-Dec processes the output of the $r$-th layer in T-Enc, where $r$ is a hyperparameter. The auxiliary CM loss can be denoted as follows.

\begin{equation}
    \mathcal{L}_{\mathrm{CM}} = - \mathrm{log} P(\widetilde{u_s}| a_{s}, \theta_{\mathrm{A\text{-}Enc}}, \theta_{\mathrm{T\text{-}Enc}}^r,\theta_{\mathrm{S\text{-}Dec}})
\end{equation}
where $\theta_{\mathrm{T\text{-}Enc}}^r$ indicates that the parameters before the $r$-th layer in T-Enc are used.
Similarly, the target unit language $\widetilde{u_t}$ can be generated according to $u_t$. We view $\widetilde{u_t}$ as translation text and implement the T-Dec to compute the auxiliary CL loss:
\begin{equation}
    \mathcal{L}_{\mathrm{CL}} = - \mathrm{log} P(\widetilde{u_t}| a_{s}, \theta_{\mathrm{A\text{-}Enc}}, \theta_{\mathrm{T\text{-}Enc}}, \theta_{\mathrm{T\text{-}Dec}}).
\end{equation}
We combine all the training losses to form the final training objective:
\begin{equation}
    \mathcal{L} = \mathcal{L}_{\mathrm{TU}} + \alpha \mathcal{L}_{\mathrm{SU}} + \beta \mathcal{L}_{\mathrm{CM}} + \gamma \mathcal{L}_{\mathrm{CL}}
\end{equation}
where $\alpha$ is fixed at 8. $\beta$ and $\gamma$ are set to 8 if the corresponding task is activated. We also test the multi-task learning approach by applying text as the auxiliary data. The losses $\mathcal{L}_{\mathrm{CM}^{'}}$ and $\mathcal{L}_{\mathrm{CL}^{'}}$ represent the scenarios where we replace the unit language with the source and target text, respectively.
Table \ref{MTL} provides brief information about terms of multi-task learning.

\begin{figure}[t]
\resizebox{.48\textwidth}{!}{
    \centering
    \small
      \begin{tikzpicture} [scale=1.0]

        \node(te) at (0,0) [rectangle, draw=black, fill=orange!10, rounded corners=3pt, thick, minimum width=2.8cm,minimum height=2.8cm,align=center,anchor=south] {};

        \node at ([xshift=-2cm, yshift=0.0cm]te.north) [rectangle, fill=none, rounded corners=3pt, thick, minimum width=1.8cm,minimum height=0.7cm,align=center,anchor=north] {Textual\\encoder};

        \node[rotate = 180] at ([yshift=-0.075cm ]te.south) {$\underbrace{\hspace{2.1cm}}$};
        \node[rotate = 180] at ([yshift=0.5cm]te.center) {$\underbrace{\hspace{2.1cm}}$};
        
        \draw[->,thick]([yshift=0cm]te.south)--([yshift=0.2cm]te.south);
        \draw[->,thick]([yshift=-0.7cm]te.center)--([yshift=-0.365cm]te.center);
        \draw[->,thick]([yshift=-0.3cm]te.north)--([yshift=0.2cm]te.north);

        \node(layers2) at ([xshift=0.3cm, yshift=-0.75cm] te.north) [rectangle, draw=black, fill=black!10, rounded corners=3pt, thick, minimum width=1.8cm,minimum height=0.6cm,align=center,anchor=south] {Layers};
        \node(tgt_decoder) at ([xshift=0cm, yshift=1.2cm]layers2.north) [rectangle, fill=none, rounded corners=3pt, thick, minimum width=1.8cm,minimum height=0.7cm,align=center,anchor=south] {Target unit decoder};

        \node[rotate = 180] at ([xshift=0.1cm,yshift=0cm]tgt_decoder.south) {$\underbrace{\hspace{1.4cm}}$};

        \draw[->,thick]([xshift=0.1cm,yshift=0.05cm]tgt_decoder.south)--([xshift=0.1cm,yshift=0.25cm]tgt_decoder.south);

        \node(layers1) at ([xshift=0.3cm, yshift=0.2cm]te.south) [rectangle, draw=black, fill=black!10, rounded corners=3pt, thick, minimum width=1.8cm,minimum height=0.6cm,align=center,anchor=south] {Layers};

        \node(feat_in) at ([xshift=-0.3cm, yshift=-0.9cm]te.south) [rectangle, draw=black, fill=red!10, rounded corners=0.5pt, thick, minimum width=0.3cm,minimum height=0.7cm,align=center,anchor=south] {};
        \node at ([xshift=0.45cm]feat_in.south) [rectangle, draw=black, fill=red!10, rounded corners=0.5pt, thick, minimum width=0.3cm,minimum height=0.7cm,align=center,anchor=south] {};
        \node at ([xshift=0.9cm]feat_in.south) [rectangle, fill=none, draw=none, rounded corners=0.5pt, thick, minimum width=0.3cm, minimum height=0.7cm, align=center, anchor=south] {···};
        \node at ([xshift=1.35cm]feat_in.south) [rectangle, draw=black, fill=red!10, rounded corners=0.5pt, thick, minimum width=0.3cm,minimum height=0.7cm,align=center,anchor=south] {};

        \node at ([xshift=0.675cm, yshift=-0.6cm ]feat_in.south)[rectangle, fill=none, rounded corners=0.5pt, thick, minimum width=0.3cm,minimum height=0.7cm,align=center,anchor=south] {Features};
        
        \node(bcm) at ([xshift=-0.75cm]feat_in.south) [rectangle, draw=black, fill=green!10, rounded corners=0.5pt, thick, minimum width=0.3cm,minimum height=0.7cm,align=center,anchor=south] {};

        \node at ([xshift=0cm, yshift=-0.6cm ]bcm.south)[rectangle, fill=none, rounded corners=0.5pt, thick, minimum width=0.3cm,minimum height=0.7cm,align=center,anchor=south] {$b_{CM}$};

        \node(feat_tmp) at ([xshift=-0.3cm, yshift=-0.35cm]te.center) [rectangle, draw=black, fill=red!10, rounded corners=0.5pt, thick, minimum width=0.3cm, minimum height=0.7cm, align=center, anchor=south] {};
        \node at ([xshift=0.45cm]feat_tmp.south) [rectangle, draw=black, fill=red!10, rounded corners=0.5pt, thick, minimum width=0.3cm,minimum height=0.7cm,align=center,anchor=south] {};
        \node at ([xshift=0.9cm]feat_tmp.south) [rectangle, fill=none, draw=none, rounded corners=0.5pt, thick, minimum width=0.3cm, minimum height=0.7cm, align=center, anchor=south] {···};
        \node at ([xshift=1.35cm]feat_tmp.south) [rectangle, draw=black, fill=red!10, rounded corners=0.5pt, thick, minimum width=0.3cm,minimum height=0.7cm,align=center,anchor=south] {};

        \node(org_bcm) at ([xshift=-0.75cm]feat_tmp.south) [rectangle, draw=black, fill=green!10, rounded corners=0.5pt, thick, minimum width=0.3cm,minimum height=0.7cm,align=center,anchor=south] {};

        \node (bcl) at ([xshift=-2.5cm]feat_tmp.south) [rectangle, draw=black, fill=blue!10, rounded corners=0.5pt, thick, minimum width=0.3cm,minimum height=0.7cm,align=center,anchor=south] {};
        \node at ([xshift=0cm, yshift=-0.55cm ]bcl.south)[rectangle, fill=none, rounded corners=0.5pt, thick, minimum width=0.3cm,minimum height=0.7cm,align=center,anchor=south] {$b_{CL}$};
        \node at ([xshift=0.6cm, yshift=-1.2cm]bcl.east) [rectangle, fill=none, rounded corners=0.5pt, thick, minimum width=0.3cm,minimum height=0.7cm,align=center,anchor=south] {Replace\\at $r_{th}$\\layer};

        \draw[->,thick, dashed](bcl.east)--(org_bcm.west);

        \node(feat_out) at ([xshift=-0.3cm, yshift=0.25cm]te.north) [rectangle, draw=black, fill=red!10, rounded corners=0.5pt, thick, minimum width=0.3cm, minimum height=0.7cm, align=center, anchor=south] {};
        \node at ([xshift=0.45cm]feat_out.south) [rectangle, draw=black, fill=red!10, rounded corners=0.5pt, thick, minimum width=0.3cm,minimum height=0.7cm,align=center,anchor=south] {};
        \node at ([xshift=0.9cm]feat_out.south) [rectangle, fill=none, draw=none, rounded corners=0.5pt, thick, minimum width=0.3cm, minimum height=0.7cm, align=center, anchor=south] {···};
        \node at ([xshift=1.35cm]feat_out.south) [rectangle, draw=black, fill=red!10, rounded corners=0.5pt, thick, minimum width=0.3cm,minimum height=0.7cm,align=center,anchor=south] {};

        \node at ([xshift=-1.35cm, yshift=-0.1cm]feat_out.west) [rectangle, fill=none, rounded corners=3pt, thick, minimum width=1.8cm,minimum height=0.7cm,align=center,anchor=south] {Drop};
        \draw[->,thick, dashed]([xshift=-0.76cm]feat_out.west)--([xshift=-1.6cm]feat_out.west);

        \node  at ([xshift=-0.75cm]feat_out.south) [rectangle, draw=black, fill=blue!10, rounded corners=0.5pt, thick, minimum width=0.3cm,minimum height=0.7cm,align=center,anchor=south] {};

        \node(sd) at ([xshift=2cm,]layers1.east) [rectangle, draw=black, fill=lblue, rounded corners=3pt, thick, minimum width=2.4cm,minimum height=1cm,align=center,anchor=center] {Source text\\decoder};
        \draw[->,thick]([yshift=-0.5cm]te.center)--([xshift=1.5cm,yshift=-0.5cm]te.center)--([xshift=1.5cm]te.center|-sd.west)--(sd.west);
        \draw[->,thick](sd.north)--([yshift=0.4cm]sd.north);
        \node(loss_cm) at ([yshift=0.2cm]sd.north) [rectangle, fill=none, rounded corners=0.5pt, thick, minimum width=0.3cm,minimum height=0.7cm,align=center,anchor=south] {$\mathcal{L}_{CM}$};

        \node(td) at ([xshift=2cm]layers2.east) [rectangle, draw=black, fill=lblue, rounded corners=3pt, thick, minimum width=2.4cm,minimum height=1cm,align=center,anchor=center] {Target text\\decoder};
        \draw[->,thick]([yshift=0.05cm]te.north)--([xshift=1.5cm,yshift=0.05cm]te.north)--([xshift=1.5cm]te.north|-td.west)--(td.west);
        \draw[->,thick](td.north)--([yshift=0.4cm]td.north);

        \node(loss_cl) at ([yshift=0.2cm]td.north) [rectangle, fill=none, rounded corners=0.5pt, thick, minimum width=0.3cm,minimum height=0.7cm,align=center,anchor=south] {$\mathcal{L}_{CL}$};

        \node(mse) at ([xshift=-2.5cm, yshift=0cm]feat_in.south) [rectangle, fill=none, rounded corners=0.5pt, thick, minimum width=0.3cm,minimum height=0.7cm,align=center,anchor=south] {$\mathcal{L}_{MSE}$};
        \draw[->,thick]([yshift=-0.4cm]bcl.south)--([yshift=-0.2cm]mse.north);
        \draw[->,thick]([xshift=-0.1cm]bcm.west)--(mse.east);

        
      \end{tikzpicture}
      }
      \caption{The guiding process of the task prompt.}
      \label{task_prrompt}
      \vspace{-0.2cm}
  \end{figure}

  
\subsection{Task Prompt Modeling}
In our subsequent analysis, we observe a conflict between $\mathcal{L}_{\mathrm{CM}}$ and $\mathcal{L}_{\mathrm{CL}}$.  To address this issue, we introduce task prompts to improve multi-task learning when applying the two tasks simultaneously. Specifically, we employ two learnable weights as task prompts, namely $b_{\mathrm{CM}} \in \mathbb{R}^{1 \times h}$ and $b_{\mathrm{CL}} \in \mathbb{R}^{1 \times h}$, where $h$ is the hidden size.

\begin{table*}[t]
    \centering
    \small
    \setlength{\tabcolsep}{1.5mm}{
    \begin{tabular}{lllll}
    \toprule
     we are human beings.    \\ 
     \midrule
     {\color{gray}704\_334 365\_548\_985 99\_991} {\color{blue}535\_271\_930} (we) {\color{blue}327\_905\_579} (are) {\color{gray}933\_901} \\{\color{blue}427\_258\_436 139\_340\_748 872\_336\_877} (human) {\color{gray}488\_620\_915}\\ {\color{blue}143\_290\_978  485\_113} (be) {\color{blue}398\_212\_455} {\color{blue}545\_711\_510} (ings) {\color{gray}337\_243 59} \\
    \midrule
    we are relatives.     \\ 
    \midrule
    {\color{gray}681\_63 991\_162} {\color{blue}535\_271\_930} (we) {\color{blue}327\_905\_579} (are) {\color{gray}969\_156\_824} \\ {\color{blue}384\_879 259\_317\_453 275\_830\_471 737\_53\_885 545\_85\_510} (relatives) {\color{gray}297\_206\_265} \\
    \bottomrule
    \end{tabular}}
    \caption{Comparison between unit language and text. The gray unit language denotes the noise token.}
    \label{Case_study}
\end{table*}

We use $b_{\mathrm{CM}}$ and $b_{\mathrm{CL}}$ as inductive biases for the CM and CL tasks respectively. The overall processing is shown in Figure \ref{task_prrompt}. Before the features are fed into the T-Enc, the prompt $b_{\mathrm{CM}}$ is concatenated with the features at the first position. The intermediate features of the $r$-th layer are used to compute $\mathcal{L}_{\mathrm{CM}}$. Then the task prompt $\mathcal{L}_{\mathrm{CM}}$ is replaced by $b_{\mathrm{CL}}$ after the $r$-th layer to adapt the CL modeling. Furthermore, we add an extra loss, calculated by the mean square error between $b_{\mathrm{CM}}$ and $b_{\mathrm{CL}}$ with a negative weight of -3.0. This enhances the diversity of the prompts and leads to improved performance.

\section{Experiments}
\subsection{Data and Model Settings}

We conducted experiments on the VoxPopuli speech-to-speech dataset \cite{wang-etal-2021-voxpopuli}. The source and target units are converted by mHubert \cite{lee-etal-2022-textless} with 1000 units. The input speech features are 80-dimensional filter banks. The normalized unit is generated by the speaker normalizer \cite{lee-etal-2022-textless}. We set $K$ to 3 to search for the unit language for all translation pairs. We use SentencePiece \cite{kudo2018sentencepiece} to control the vocabulary size to 10k. For S2ST training with text, we utilize pre-trained ASR models\footnote{En: https://huggingface.co/facebook/wav2vec2-large-960h-lv60-self, Es: https://huggingface.co/jonatasgrosman/wa\\v2vec2-large-xlsr-53-spanish, Fr: https://huggingface.co/jona\\tasgrosman/wav2vec2-large-fr-voxpopuli-french} and split the words to the character level as the training target.   

We use the Transformer \cite{vaswani2017attention} as the backbone network with a 12-layer encoder and a 6-layer decoder. The hidden size is 512. The unit representation is the output of the 6th layer. The source and target text decoders are set to 2 layers. The parameter $r$ is set to 2, meaning the source decoders use the output of the 2nd T-Enc layer. The target decoders are implemented after the T-Enc, which is the 12th layer of the whole encoder. The vocoder we used is the unit-based HiFi-GAN \cite{kong2020hifi, polyak2021speech}.

During the training stage, we adopt a different strategy, utilizing a larger learning rate (0.001) and a bigger batch size (40,000 tokens). This approach enables our models to converge faster and achieve better performance. All experiments are conducted on 8 RTX 3090 GPUs. We use the best checkpoint for evaluation. Pre-trained ASR models$^{1}$ are used to recognize speech generated by the S2ST model. To normalize references, we remove punctuation, convert numbers to spoken forms, and lowercase text, following the work of \citet{lee-etal-2022-textless}. We report ASR SacreBLEU \cite{post-2018-call}. More training and data details can be found in the Appendix.


\usetikzlibrary{pgfplots.groupplots}

\begin{figure}[t]
    \begin{minipage}[b]{0.3\linewidth}
    \subfigure[En-Es]{
    \centering
    \begin{tikzpicture}[xscale=.75, yscale=.75, font=\fontsize{10}{10}\selectfont]
    \scriptsize{
    \begin{groupplot}[
    group style={
        group name=my fancy plots,
        group size=1 by 2,
        xticklabels at=edge bottom,
        vertical sep=0pt,
    },
    ybar,
    width=2.35\linewidth,
    height=1.4\linewidth,
    xmin=0., xmax=6,
    legend style={
        at={(0.6,1.55)},
        anchor=south,
        legend columns=-1
    },
    axis x line=bottom,
    xtick={0.6,1.85,3,4.35,5.6},
    xticklabels={Frame, Unit, Char, Unit\_l, Text},
    ymajorgrids, xmajorgrids, grid style=dashed,
]

    \nextgroupplot[ymin=1000,ymax=1250,
               ytick={1100,1200},
               axis x line=top,
               axis y discontinuity=crunch,
               axis line style={-}
               ]

    \addplot[fill=blue!30, draw=blue,area legend,ybar,bar width=0.6em,bar shift=-0.37em,enlarge x limits=0.2] coordinates {({0.6},1182.441891)};
    \addplot[fill=red!30, draw=red,area legend,ybar,bar width=0.6em,bar shift=0.37em] coordinates {({0.6},1197.166186)};
 
    \nextgroupplot[ymin=0,ymax=400,
               ytick={100,200,300,400},
               axis x line=bottom,
               legend entries={En\ \ ,Es\ \ },
               ylabel={Avg. tokens},
               xlabel style={yshift=5pt}, 
               ylabel style={xshift=14pt, yshift=-10pt}, 
               axis line style={-}
               ]
    \addplot[fill=blue!30, draw=blue,area legend,ybar,bar width=0.6em,bar shift=-0.37em] coordinates {({0.6},500) ({1.85},358.346953) ({3},141.3772385) ({4.35},130.5482898) ({5.6},30.41168685)};
    \addplot[fill=red!30, draw=red,area legend,ybar,bar width=0.6em,bar shift=0.37em] coordinates {({0.6},500) ({1.85},309.7904781) ({3},156.7483401) ({4.35},110.907598) ({5.6},32.00630703)};
\end{groupplot}
    }
    \draw[color=blue!30, line width=2pt] (0.28,0) -- (0.41,0);
    \draw[color=red!30, line width=2pt] (0.46,0) -- (0.59,0);
  \end{tikzpicture}
  }
  \end{minipage}
  \hspace{40pt}
  \begin{minipage}[b]{0.3\linewidth}
  \subfigure[En-Fr]{
    \begin{tikzpicture}[xscale=.75, yscale=.75,font=\fontsize{10}{10}\selectfont]
    \scriptsize{
    \begin{groupplot}[
    group style={
        group name=my fancy plots,
        group size=1 by 2,
        xticklabels at=edge bottom,
        vertical sep=0pt,
    },
    ybar,
    width=2.35\linewidth,
    height=1.4\linewidth,
    xmin=0, xmax=6,
    legend style={
        at={(0.6,1.55)},
        anchor=south,
        legend columns=-1
    },
    xtick={0.6,1.85,3,4.35,5.6},
    xticklabels={Frame, Unit, Char, Unit\_l, Text},
    ymajorgrids, xmajorgrids, grid style=dashed,
]

\nextgroupplot[ymin=1000,ymax=1250,
               ytick={1100,1200},
               axis x line=top,
               axis y discontinuity=crunch,
               axis line style={-}
               ]

\addplot[fill=blue!30, draw=blue,area legend,ybar,bar width=0.6em,bar shift=-0.37em,enlarge x limits=0.2] coordinates {({0.6},1187.135381)};
\addplot[fill=red!30, draw=red,area legend,ybar,bar width=0.6em,bar shift=0.37em] coordinates {({0.6},1214.616934)};

\nextgroupplot[ymin=0,ymax=400,
    xtick={0.6,1.85,3,4.35,5.6},
    xticklabels={Frame, Unit, Char, Unit\_l, Text},
               ytick={100,200,300,400},
               axis x line=bottom,
               legend entries={En\ \ ,Fr\ \ },
               xlabel style={yshift=5pt}, 
               ylabel style={xshift=14pt, yshift=-10pt}, 
               axis line style={-}
               ]

 \addplot[fill=blue!30, draw=blue,area legend,ybar,bar width=0.6em,bar shift=-0.37em,enlarge x limits=0.15] coordinates {({0.6},1187.135381) ({1.85},360.4508883) ({3},143.0095647) ({4.35},131.2596106) ({5.6},30.8449947)};
\addplot[fill=red!30, draw=red,area legend,ybar,bar width=0.6em,bar shift=0.37em] coordinates {({0.6},1214.616934) ({1.85},319.7755849) ({3},147.1623683) ({4.35},114.9738985) ({5.6},29.49783419)  };

\end{groupplot}
\draw[color=blue!30, line width=2pt] (0.28,0) -- (0.41,0);
\draw[color=red!30, line width=2pt] (0.46,0) -- (0.59,0);
    }
  \end{tikzpicture}
  }
    \end{minipage}
    \captionsetup{skip=2pt}
    \caption{Average lengths of different types of tokens on En-Es and En-Fr training datasets. ``Unit\_l'' denotes the unit language. Note that the units used here have had continuous repetitions removed.}
    \label{Length_ratio}
\end{figure}
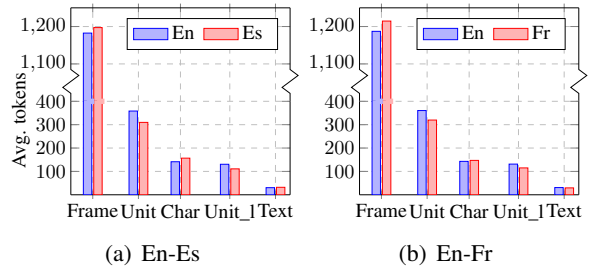

\begin{figure*}[t]
    \centering
    \subfigure{
    \begin{minipage}[t]{0.485\linewidth}
    \centering
    \begin{tikzpicture}
  
        \pgfplotsset{set layers}
         \scriptsize{
         \begin{axis}
        [
       at={(0,0)},
          ymajorgrids,
          xmajorgrids,
          grid style=dashed,
          width=1.\textwidth,
          height=.48\textwidth,
          legend style={at={(0.3,0.78)}, anchor=south west},
          ylabel={\scriptsize{Sparseness}},
          ylabel style={yshift=-1.5em},
          yticklabel style={/pgf/number format/fixed zerofill, /pgf/number format/fixed,/pgf/number format/sci precision=2},
          ymin=0.0018,ymax=0.005, ytick={0.002,0.0025,0.003,0.0035,0.004,0.0045,0.0045},
          xmin=0,xmax=13,xtick={1,2,3,4,5,6,7,8,9,10,11,12},
          legend columns=4,
          legend style={yshift=-1pt,xshift=1.5em, legend plot pos=left,cells={anchor=west}}
          ]
          \addplot[blue!60,mark=pentagon*,mark size=1.5pt,thick,mark options={fill=white,draw=blue,line width=0.5pt}] coordinates {
(1 ,0.00355349)
(2 ,0.003160276)
(3 ,0.00283684)
(4 ,0.002616306)
(5 ,0.002430006)
(6 ,0.002538524)
(7 ,0.002423785)
(8 ,0.002410502)
(9 ,0.00235246)
(10,0.002355352)
(11,0.002498877)
(12,0.003282918)}; 
          \addlegendentry{\scalebox{.8}{Baseline}}
          \addplot[teal!70,mark=diamond*,mark size=1.5pt,thick,mark options={fill=white,draw=teal,line width=0.5pt}] coordinates {
(1 ,0.004205541)
(2 ,0.003815925)
(3 ,0.003413496)
(4 ,0.003022098)
(5 ,0.002821001)
(6 ,0.002821237)
(7 ,0.003014498)
(8 ,0.002436228)
(9 ,0.002386693)
(10,0.002514245)
(11,0.002807315)
(12,0.003588968)}; 

          \addlegendentry{\scalebox{.8}{CM}}
          \addplot[orange!80,mark=triangle*,,mark size=1.5pt,thick,mark options={fill=white,draw=orange,line width=0.5pt}] coordinates {
(1 ,0.003774764)
(2 ,0.003547538)
(3 ,0.002966813)
(4 ,0.002835596)
(5 ,0.002560719)
(6 ,0.002575213)
(7 ,0.002581905)
(8 ,0.002408148)
(9 ,0.00236618)
(10,0.0022164)
(11,0.002195013)
(12,0.002197804)};

           \addlegendentry{\scalebox{.8}{CL}}

          \end{axis}}

        \end{tikzpicture}
    \end{minipage}
    }
    \subfigure{
    \begin{minipage}[t]{0.485\linewidth}
    \centering
\begin{tikzpicture}
        \pgfplotsset{set layers}
         \scriptsize{
         \begin{axis}
        [
       at={(0,0)},
          ymajorgrids,
          xmajorgrids,
          grid style=dashed,
          width=1\textwidth,
          height=.48\textwidth,
          legend style={at={(0.3,0.78)}, anchor=south west},
          ylabel={\scriptsize{Sparseness}},
          ylabel style={yshift=-1.5em},
          yticklabel style={/pgf/number format/fixed zerofill, /pgf/number format/fixed,/pgf/number format/sci precision=2},
          ymin=0.0018,ymax=0.0055, ytick={0.002,0.0025,0.003,0.0035,0.004,0.0045,0.005,0.005},
          xmin=0,xmax=13,xtick={1,2,3,4,5,6,7,8,9,10,11,12},
          legend columns=4,
          legend style={yshift=-1pt,xshift=1.5em, legend plot pos=left,cells={anchor=west}}
          ]
          \addplot[blue!60,mark=pentagon*,mark size=1.5pt,thick,mark options={fill=white,draw=blue,line width=0.5pt}] coordinates {
(1 ,0.003446702)
(2 ,0.002953326)
(3 ,0.002681769)
(4 ,0.002446278)
(5 ,0.002326999)
(6 ,0.002452762)
(7 ,0.002414046)
(8 ,0.002290263)
(9 ,0.002242666)
(10,0.00220938)
(11,0.002504735)
(12,0.003588247)};

          \addlegendentry{\scalebox{.8}{Baseline}}
          \addplot[teal!70,mark=diamond*,mark size=1.5pt,thick,mark options={fill=white,draw=teal,line width=0.5pt}] coordinates {
(1 ,0.004158193)
(2 ,0.003717716)
(3 ,0.003191214)
(4 ,0.002866087)
(5 ,0.002704354)
(6 ,0.002678224)
(7 ,0.002890269)
(8 ,0.002492054)
(9 ,0.002326615)
(10,0.002480298)
(11,0.002765719)
(12,0.003826518)}; 
        
          \addlegendentry{\scalebox{.8}{CM}}
          \addplot[orange!80,mark=triangle*,,mark size=1.5pt,thick,mark options={fill=white,draw=orange,line width=0.5pt}] coordinates {
(1 ,0.004463514)
(2 ,0.003877404)
(3 ,0.003620382)
(4 ,0.003498132)
(5 ,0.003129689)
(6 ,0.00302421)
(7 ,0.002993991)
(8 ,0.00282063)
(9 ,0.00274799)
(10,0.002589227)
(11,0.002743549)
(12,0.002672538)};

          \addlegendentry{\scalebox{.8}{CL}}

          \end{axis}}

        \end{tikzpicture}
    \end{minipage}
    }
    \captionsetup{skip=2pt}
    \caption{The influences of CM and CL on Fr-En (left) and En-Fr (right) tasks.}
    \label{unit_language_influence}
    \vspace{-0.2cm}
\end{figure*}

\begin{figure*}[t]
    \centering
    \subfigure{
    \begin{minipage}[t]{0.485\linewidth}
    \centering
    \begin{tikzpicture}
  
        \pgfplotsset{set layers}
         \scriptsize{
         \begin{axis}
        [
       at={(0,0)},
          ymajorgrids,
          xmajorgrids,
          grid style=dashed,
          width=1.\textwidth,
          height=.48\textwidth,
          legend style={at={(0.28,0.74)}, anchor=south west},
          ylabel={\scriptsize{Sparseness}},
          ylabel style={yshift=-1.5em},
          yticklabel style={/pgf/number format/fixed zerofill, /pgf/number format/fixed,/pgf/number format/sci precision=2},
          ymin=0.002,ymax=0.0055, ytick={0.0025,0.003,0.0035,0.004,0.0045,0.005},
          xmin=0,xmax=13,xtick={1,2,3,4,5,6,7,8,9,10,11,12},
          legend columns=4,
          legend style={yshift=-1pt,xshift=1.5em, legend plot pos=left,cells={anchor=west}}
          ]
          \addplot[blue!60,mark=pentagon*,mark size=1.5pt,thick,mark options={fill=white,draw=blue,line width=0.5pt}] coordinates {
(1 ,0.00355349)
(2 ,0.003160276)
(3 ,0.00283684)
(4 ,0.002616306)
(5 ,0.002430006)
(6 ,0.002538524)
(7 ,0.002423785)
(8 ,0.002410502)
(9 ,0.00235246)
(10,0.002355352)
(11,0.002498877)
(12,0.003282918)}; 

          \addlegendentry{\scalebox{.8}{Baseline}}
          \addplot[teal!70,mark=diamond*,mark size=1.5pt,thick,mark options={fill=white,draw=teal,line width=0.5pt}] coordinates {
(1 ,0.004648121)
(2 ,0.003805668)
(3 ,0.003516432)
(4 ,0.00304436)
(5 ,0.002796991)
(6 ,0.002469049)
(7 ,0.002506208)
(8 ,0.002387803)
(9 ,0.002269398)
(10,0.00239167)
(11,0.002462054)
(12,0.003292872)}; 
     
          \addlegendentry{\scalebox{.8}{CM$^{'}$}}
          \addplot[orange!80,mark=triangle*,,mark size=1.5pt,thick,mark options={fill=white,draw=orange,line width=0.5pt}] coordinates {
(1 ,0.004965202)
(2 ,0.004376944)
(3 ,0.003997281)
(4 ,0.003827291)
(5 ,0.003295159)
(6 ,0.003085016)
(7 ,0.00312315)
(8 ,0.002913311)
(9 ,0.002743925)
(10,0.002538255)
(11,0.002675963)
(12,0.002559979)};

           \addlegendentry{\scalebox{.8}{CL$^{'}$}}

          \end{axis}}

        \end{tikzpicture}
    \end{minipage}
    }
    \subfigure{
    \begin{minipage}[t]{0.485\linewidth}
    \centering
\begin{tikzpicture}
        \pgfplotsset{set layers}
         \scriptsize{
         \begin{axis}
        [
       at={(0,0)},
          ymajorgrids,
          xmajorgrids,
          grid style=dashed,
          width=1\textwidth,
          height=.48\textwidth,
          legend style={at={(0.28,0.74)}, anchor=south west},
          ylabel={\scriptsize{Sparseness}},
          ylabel style={yshift=-1.5em},
          yticklabel style={/pgf/number format/fixed zerofill, /pgf/number format/fixed,/pgf/number format/sci precision=2},
          ymin=0.0020,ymax=0.0060, ytick={0.0025,0.003,0.0035,0.004,0.0045,0.005,0.0055},
          xmin=0,xmax=13,xtick={1,2,3,4,5,6,7,8,9,10,11,12},
          legend columns=4,
          legend style={yshift=-1pt,xshift=1.5em, legend plot pos=left,cells={anchor=west}}
          ]
          \addplot[blue!60,mark=pentagon*,mark size=1.5pt,thick,mark options={fill=white,draw=blue,line width=0.5pt}] coordinates {
(1 ,0.003446702)
(2 ,0.002953326)
(3 ,0.002681769)
(4 ,0.002446278)
(5 ,0.002326999)
(6 ,0.002452762)
(7 ,0.002414046)
(8 ,0.002290263)
(9 ,0.002242666)
(10,0.00220938)
(11,0.002504735)
(12,0.003588247)};

          \addlegendentry{\scalebox{.8}{Baseline}}
          \addplot[teal!70,mark=diamond*,mark size=1.5pt,thick,mark options={fill=white,draw=teal,line width=0.5pt}] coordinates {
(1 ,0.003938801)
(2 ,0.003412618)
(3 ,0.003176935)
(4 ,0.002781307)
(5 ,0.00252601)
(6 ,0.002314221)
(7 ,0.002405581)
(8 ,0.00219574)
(9 ,0.002089526)
(10,0.002231965)
(11,0.002364916)
(12,0.003369462)}; 
        
          \addlegendentry{\scalebox{.8}{CM$^{'}$}}
          \addplot[orange!80,mark=triangle*,,mark size=1.5pt,thick,mark options={fill=white,draw=orange,line width=0.5pt}] coordinates {
(1 ,0.005380565)
(2 ,0.005103131)
(3 ,0.004312515)
(4 ,0.003869993)
(5 ,0.003612396)
(6 ,0.003306276)
(7 ,0.003205589)
(8 ,0.003080623)
(9 ,0.002885221)
(10,0.002743422)
(11,0.002800761)
(12,0.0029064)};

          \addlegendentry{\scalebox{.8}{CL$^{'}$}}

          \end{axis}}

        \end{tikzpicture}
    \end{minipage}
    }
    \captionsetup{skip=2pt}
    \caption{The influences of CM$^{'}$ and CL$^{'}$ on Fr-En (left) and En-Fr (right) tasks.}
    \label{text_influence}
\end{figure*}

\subsection{Results}

We compare the advanced textless S2ST with the text-based model \cite{barrault2023seamlessm4t} in Table \ref{main}. This comparison shows that the textless model has significant improvement potential. After applying unit language as the auxiliary training data, results in Table \ref{main} show that both $\mathcal{L}_{\mathrm{CL}}$ and $\mathcal{L}_{\mathrm{CM}}$ improve the performance of the textless model. The two losses achieve average improvements of 0.4 to 0.8 BLEU based on strong baselines. All the translation pairs show consistent improvement. We find that the improvement gained from CL training is greater than that from CM training, which indicates that the textless S2ST requires crucial enhancement in cross-lingual learning.

We further compare our method with a text-based method. The results also prove that both source and target texts have much potential to improve the S2UT. We find that our method shows almost comparable performance with the text-based method in cross-lingual learning. This demonstrates that our unit language, mined through language modeling, can effectively function like text.

When we apply the $\mathcal{L}_{\mathrm{CL}^{'}}$ and $\mathcal{L}_{\mathrm{CM}^{'}}$ methods simultaneously, the performance does not further increase compared with either method alone, and even degradation occurs. The Fr-En and En-Es pairs trained with the unit language confirm this phenomenon. This indicates that the effects of the two methods may not be the same and that conflicts have occurred during the modeling process. We aim to isolate the effects of $\mathcal{L}_{\mathrm{CL}}$ and $\mathcal{L}_{\mathrm{CM}}$ by adding task prompts method. Results in Table \ref{main} demonstrate that this approach harmonizes CM and CL methods, achieving much advanced performance in all textless S2ST directions. Furthermore, our unit language can help the model achieve performance comparable to true text, which could significantly aid languages without sufficient labeled text for modeling speech.

\section{Analysis}

\subsection{Comparison of Unit Language and Text}
Since the unit sequence is rather long, we choose some short speech samples for a case study to compare with text. As shown in Table \ref{Case_study}, the unit language is basically merged according to pronunciation. Thus, the size of the unit language representing a word depends on the syllables of the word. Furthermore, even across different sentences, the unit language can successfully represent the same word. Concise sequences benefit the extraction of important information \cite{chan2015listen}, thus the unit language could address the CL and CM modeling challenges in textless S2ST.

We present the average length of different types of training data in Figure \ref{Length_ratio}. It is very difficult to learn cross-lingual alignment based on thousands of various frames if there is no intermediate guidance. Units could help with modeling speech and they are several times shorter than frames \cite{lee-etal-2022-textless}. But units still have an obvious inconsistency in length compared with characters or text as shown in Figure \ref{Length_ratio}. Previous work suggests that length reduction benefits cross-lingual modeling in speech translation \cite{zhang-etal-2023-rethinking}. After applying language modeling, we find that the length of the unit language is between that of characters and text, making it suitable for learning cross-lingual alignment. Furthermore, we found that unit language is stable across different languages and consistently shows significant compression of speech sequences. This demonstrates that unit language has great potential for speech-related tasks, which could boost CM and CL learning.

\begin{figure*}[t]
    \centering
    \subfigure[Es-En]{
    \begin{minipage}[t]{0.232\linewidth}
    \centering
    \begin{tikzpicture}
        \pgfplotsset{set layers}
         \scriptsize{
         \begin{axis}
        [
       at={(0,0)},
          ymajorgrids,
          xmajorgrids,
          grid style=dashed,
          width=1\textwidth,
          height=1\textwidth,
          legend style={at={(1.8,1.2)}, anchor=south west},
          ylabel={\scriptsize{Localness}},
          ylabel style={yshift=-1.5em},
          yticklabel style={/pgf/number format/fixed zerofill, /pgf/number format/fixed,/pgf/number format/sci precision=2},
          ymin=0.07,ymax=0.73, ytick={0.2,0.3,0.4,0.5,0.6},
          xmin=1,xmax=13,xtick={2,4,6,8,10,12},
          legend columns=4,
          legend style={yshift=-9pt,xshift=4.em, legend plot pos=left,cells={anchor=west}}
          ]
          \addplot[blue!60,mark=pentagon*,mark size=1.5pt,thick,mark options={fill=white,draw=blue,line width=0.5pt}] coordinates {

(2 ,0.36443566)
(4 ,0.429218677)
(6 ,0.489436683)
(8 ,0.284974139)
(10,0.13402115)
(12,0.128956829)};

          \addlegendentry{\scalebox{.8}{Baseline}}
          \addplot[teal!70,mark=diamond*,mark size=1.5pt,thick,mark options={fill=white,draw=teal,line width=0.5pt}] coordinates {
(2 ,0.392479929)
(4 ,0.673779028)
(6 ,0.507809936)
(8 ,0.380404408)
(10,0.175688551)
(12,0.105880428)};

          \addlegendentry{\scalebox{.8}{CM}}
          \addplot[orange!80,mark=triangle*,,mark size=1.5pt,thick,mark options={fill=white,draw=orange,line width=0.5pt}] coordinates {
(2 ,0.408373979)
(4 ,0.580926669)
(6 ,0.447904136)
(8 ,0.251891701)
(10,0.491127588)
(12,0.212998281)};

          \addlegendentry{\scalebox{.8}{CL}}
          \addplot[red!60,mark=square*,mark size=1.2pt,thick,mark options={fill=white,draw=red,line width=0.5pt}] coordinates {
(2 ,0.297057489)
(4 ,0.482680591)
(6 ,0.423444448)
(8 ,0.542326497)
(10,0.258826626)
(12,0.184249841)};
          \addlegendentry{\scalebox{.8}{CM\&CL}}
          \end{axis}}

        \end{tikzpicture}
    \end{minipage}
    }
    \subfigure[Fr-En]{
    \begin{minipage}[t]{0.232\linewidth}
    \centering
    \begin{tikzpicture}
        \pgfplotsset{set layers}
         \scriptsize{
         \begin{axis}
        [
       at={(0,0)},
          ymajorgrids,
          xmajorgrids,
          grid style=dashed,
          width=1\textwidth,
          height=1\textwidth,
          legend style={at={(0.01,0.91)}, anchor=south west},
          ylabel={\scriptsize{Localness}},
          ylabel style={yshift=-1.5em},
          yticklabel style={/pgf/number format/fixed zerofill, /pgf/number format/fixed,/pgf/number format/sci precision=2},
          ymin=0.05,ymax=0.75, ytick={0.2,0.3,0.4,0.5,0.6},
          xmin=1,xmax=13,xtick={2,4,6,8,10,12},
          legend columns=4,
          legend style={yshift=-9pt,xshift=0.5em, legend plot pos=left,cells={anchor=west}}
          ]
          \addplot[blue!60,mark=pentagon*,mark size=1.5pt,thick,mark options={fill=white,draw=blue,line width=0.5pt}] coordinates {
(2 ,0.364596502)
(4 ,0.581390562)
(6 ,0.445637994)
(8 ,0.292704521)
(10,0.120328317)
(12,0.113100517)};

          \addplot[teal!70,mark=diamond*,mark size=1.5pt,thick,mark options={fill=white,draw=teal,line width=0.5pt}] coordinates {
(2 ,0.348537928)
(4 ,0.455518832)
(6 ,0.474022954)
(8 ,0.547072787)
(10,0.186154938)
(12,0.112842301)};

          \addplot[orange!80,mark=triangle*,,mark size=1.5pt,thick,mark options={fill=white,draw=orange,line width=0.5pt}] coordinates {
(2 ,0.31842758)
(4 ,0.364158857)
(6 ,0.490282566)
(8 ,0.319517792)
(10,0.200472064)
(12,0.194378364)};

          \addplot[red!60,mark=square*,mark size=1.2pt,thick,mark options={fill=white,draw=red,line width=0.5pt}] coordinates {
(2 ,0.460216758)
(4 ,0.220362576)
(6 ,0.592550567)
(8 ,0.352645818)
(10,0.298106131)
(12,0.207511415)};

          \end{axis}}
        \end{tikzpicture}
    \end{minipage}
    }
    \subfigure[En-Es]{
    \begin{minipage}[t]{0.232\linewidth}
    \centering
    \begin{tikzpicture}
        \pgfplotsset{set layers}
         \scriptsize{
         \begin{axis}
        [
       at={(0,0)},
          ymajorgrids,
          xmajorgrids,
          grid style=dashed,
          width=1\textwidth,
          height=1\textwidth,
          legend style={at={(0.01,0.91)}, anchor=south west},
          ylabel={\scriptsize{Localness}},
          ylabel style={yshift=-1.5em},
          yticklabel style={/pgf/number format/fixed zerofill, /pgf/number format/fixed,/pgf/number format/sci precision=2},
          ymin=0.06,ymax=0.65, ytick={0,1,0.2,0.3,0.4,0.5,0.6},
          xmin=1,xmax=13,xtick={2,4,6,8,10,12},
          legend columns=4,
          legend style={yshift=-9pt,xshift=0.5em, legend plot pos=left,cells={anchor=west}}
          ]
          \addplot[blue!60,mark=pentagon*,mark size=1.5pt,thick,mark options={fill=white,draw=blue,line width=0.5pt}] coordinates {
(2 ,0.311947189)
(4 ,0.395196321)
(6 ,0.518754282)
(8 ,0.294968784)
(10,0.190825579)
(12,0.161509152)};

          \addplot[teal!70,mark=diamond*,mark size=1.5pt,thick,mark options={fill=white,draw=teal,line width=0.5pt}] coordinates {
(2 ,0.341752888)
(4 ,0.466275229)
(6 ,0.360301513)
(8 ,0.509040027)
(10,0.189698035)
(12,0.103094176)};

          \addplot[orange!80,mark=triangle*,,mark size=1.5pt,thick,mark options={fill=white,draw=orange,line width=0.5pt}] coordinates {
(2 ,0.293510117)
(4 ,0.407836012)
(6 ,0.537332867)
(8 ,0.208663439)
(10,0.293050501)
(12,0.20351425)};

          \addplot[red!60,mark=square*,mark size=1.2pt,thick,mark options={fill=white,draw=red,line width=0.5pt}] coordinates {
(2 ,0.407740593)
(4 ,0.514456058)
(6 ,0.441628003)
(8 ,0.495885772)
(10,0.168343702)
(12,0.185886272)};

          \end{axis}}

        \end{tikzpicture}
    \end{minipage}
    }
    \subfigure[En-Fr]{
    \begin{minipage}[t]{0.232\linewidth}
    \centering
    \begin{tikzpicture}
        \pgfplotsset{set layers}
         \scriptsize{
         \begin{axis}
        [
       at={(0,0)},
          ymajorgrids,
          xmajorgrids,
          grid style=dashed,
          width=1\textwidth,
          height=1\textwidth,
          legend style={at={(0.01,0.91)}, anchor=south west},
          ylabel={\scriptsize{Localness}},
          ylabel style={yshift=-1.5em},
          yticklabel style={/pgf/number format/fixed zerofill, /pgf/number format/fixed,/pgf/number format/sci precision=2},
          ymin=0.05,ymax=0.65, ytick={0.1,0.2,0.3,0.4,0.5,0.6},
          xmin=1,xmax=13,xtick={2,4,6,8,10,12},
          legend columns=4,
          legend style={yshift=-9pt,xshift=0.5em, legend plot pos=left,cells={anchor=west}}
          ]
          \addplot[blue!60,mark=pentagon*,mark size=1.5pt,thick,mark options={fill=white,draw=blue,line width=0.5pt}] coordinates {
(2 ,0.330687577)
(4 ,0.458784008)
(6 ,0.55698619)
(8 ,0.276331353)
(10,0.156983059)
(12,0.167135616)}; 

          \addplot[teal!70,mark=diamond*,mark size=1.5pt,thick,mark options={fill=white,draw=teal,line width=0.5pt}] coordinates {
(2 ,0.329814674)
(4 ,0.330556495)
(6 ,0.400679081)
(8 ,0.546324778)
(10,0.25627841)
(12,0.15871116)};

          \addplot[orange!80,mark=triangle*,,mark size=1.5pt,thick,mark options={fill=white,draw=orange,line width=0.5pt}] coordinates {
(2 ,0.254493014)
(4 ,0.318015056)
(6 ,0.383247879)
(8 ,0.589202552)
(10,0.488833409)
(12,0.196781778)};
          \addplot[red!60,mark=square*,mark size=1.2pt,thick,mark options={fill=white,draw=red,line width=0.5pt}] coordinates {
(2 ,0.386140251)
(4 ,0.486639317)
(6 ,0.522426518)
(8 ,0.434835271)
(10,0.253609141)
(12,0.180478346)};
          \end{axis}}

        \end{tikzpicture}
    \end{minipage}
    }
    \captionsetup{skip=2pt}
    \caption{Localness of attention weight on different tasks.}
    \label{localness}
    \vspace{-0.2cm}
\end{figure*}
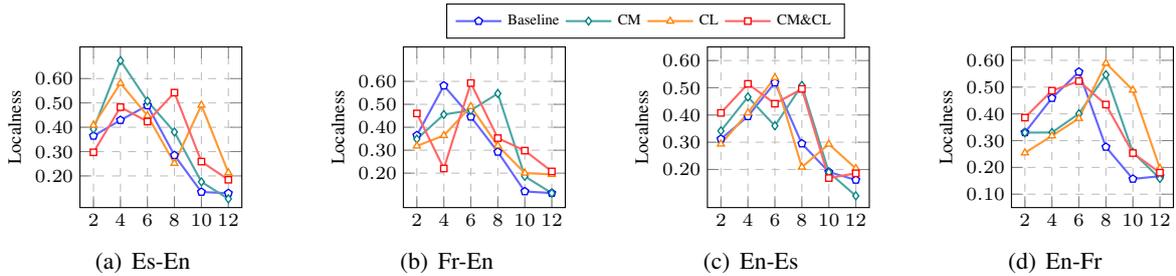

\begin{figure*}[t]
    \centering
    \subfigure[Es-En]{
    \begin{minipage}[t]{0.232\linewidth}
    \centering
    \begin{tikzpicture}
        \pgfplotsset{set layers}
         \scriptsize{
         \begin{axis}
        [
       at={(0,0)},
          ymajorgrids,
          xmajorgrids,
          grid style=dashed,
          width=1\textwidth,
          height=1\textwidth,
          legend style={at={(2.0,1.2)}, anchor=south west},
          ylabel={\scriptsize{Localness}},
          ylabel style={yshift=-1.5em},
          yticklabel style={/pgf/number format/fixed zerofill, /pgf/number format/fixed,/pgf/number format/sci precision=2},
          ymin=0.1,ymax=0.73, ytick={0.2,0.3,0.4,0.5,0.6},
          xmin=0,xmax=13,xtick={2,4,6,8,10,12},
          legend columns=4,
          legend style={yshift=-9pt,xshift=2.66em, legend plot pos=left,cells={anchor=west}}
          ]
          \addplot[red!60,mark=square*,mark size=1.2pt,thick,mark options={fill=white,draw=red,line width=0.5pt}] coordinates
          {
(1 ,0.304793848)
(2 ,0.297057489)
(3 ,0.296945586)
(4 ,0.482680591)
(5 ,0.598430813)
(6 ,0.423444448)
(7 ,0.281438761)
(8 ,0.542326497)
(9 ,0.244520256)
(10,0.258826626)
(11,0.282302709)
(12,0.184249841)};

          \addlegendentry{\scalebox{.8}{CM\&CL}}
          \addplot[blue!70,mark=diamond*,mark size=1.5pt,thick,mark options={fill=white,draw=blue,line width=0.5pt}] coordinates {
(1 ,0.304151807)
(2 ,0.229193442)
(3 ,0.526453891)
(4 ,0.316613069)
(5 ,0.381562735)
(6 ,0.598667192)
(7 ,0.391536983)
(8 ,0.404711839)
(9 ,0.24292053)
(10,0.201041084)
(11,0.224194778)
(12,0.191919727)};

          \addlegendentry{\scalebox{.8}{CM\&CL+Task prompt}}
          \end{axis}}

        \end{tikzpicture}
    \end{minipage}
    }
    \subfigure[Fr-En]{
    \begin{minipage}[t]{0.232\linewidth}
    \centering
    \begin{tikzpicture}
        \pgfplotsset{set layers}
         \scriptsize{
         \begin{axis}
        [
       at={(0,0)},
          ymajorgrids,
          xmajorgrids,
          grid style=dashed,
          width=1\textwidth,
          height=1\textwidth,
          legend style={at={(0.01,0.91)}, anchor=south west},
          ylabel={\scriptsize{Localness}},
          ylabel style={yshift=-1.5em},
          yticklabel style={/pgf/number format/fixed zerofill, /pgf/number format/fixed,/pgf/number format/sci precision=2},
          ymin=0.1,ymax=0.75, ytick={0.2,0.3,0.4,0.5,0.6},
          xmin=0,xmax=13,xtick={2,4,6,8,10,12},
          legend columns=4,
          legend style={yshift=-9pt,xshift=0.5em, legend plot pos=left,cells={anchor=west}}
          ]
          \addplot[red!60,mark=square*,mark size=1.2pt,thick,mark options={fill=white,draw=red,line width=0.5pt}] coordinates
          {
(1 ,0.193779291)
(2 ,0.460216758)
(3 ,0.474597644)
(4 ,0.220362576)
(5 ,0.368382838)
(6 ,0.592550567)
(7 ,0.440801389)
(8 ,0.352645818)
(9 ,0.261895348)
(10,0.298106131)
(11,0.235347945)
(12,0.207511415)}; 

          \addplot[blue!70,mark=diamond*,mark size=1.5pt,thick,mark options={fill=white,draw=blue,line width=0.5pt}] coordinates {
(1 ,0.196890427)
(2 ,0.528985304)
(3 ,0.361132085)
(4 ,0.383872772)
(5 ,0.44131824)
(6 ,0.463766654)
(7 ,0.304675661)
(8 ,0.317738851)
(9 ,0.229189894)
(10,0.158874513)
(11,0.152372268)
(12,0.183570458)};          
          \end{axis}}

        \end{tikzpicture}
    \end{minipage}
    }
    \subfigure[En-Es]{
    \begin{minipage}[t]{0.232\linewidth}
    \centering
    \begin{tikzpicture}
        \pgfplotsset{set layers}
         \scriptsize{
         \begin{axis}
        [
       at={(0,0)},
          ymajorgrids,
          xmajorgrids,
          grid style=dashed,
          width=1\textwidth,
          height=1\textwidth,
          legend style={at={(0.01,0.91)}, anchor=south west},
          ylabel={\scriptsize{Localness}},
          ylabel style={yshift=-1.5em},
          yticklabel style={/pgf/number format/fixed zerofill, /pgf/number format/fixed,/pgf/number format/sci precision=2},
          ymin=0.1,ymax=0.75, ytick={0.2,0.3,0.4,0.5,0.6},
          xmin=0,xmax=13,xtick={2,4,6,8,10,12},
          legend columns=4,
          legend style={yshift=-9pt,xshift=0.5em, legend plot pos=left,cells={anchor=west}}
          ]
          \addplot[red!60,mark=square*,mark size=1.2pt,thick,mark options={fill=white,draw=red,line width=0.5pt}] coordinates
          {
(1 ,0.237590456)
(2 ,0.407740593)
(3 ,0.372136467)
(4 ,0.514456058)
(5 ,0.382008756)
(6 ,0.441628003)
(7 ,0.435939686)
(8 ,0.495885772)
(9 ,0.45302876)
(10,0.168343702)
(11,0.261133185)
(12,0.185886272)};

          \addplot[blue!70,mark=diamond*,mark size=1.5pt,thick,mark options={fill=white,draw=blue,line width=0.5pt}] coordinates {
(1 ,0.199265371)
(2 ,0.370287741)
(3 ,0.489878445)
(4 ,0.201983696)
(5 ,0.422734312)
(6 ,0.415417085)
(7 ,0.268249112)
(8 ,0.409454721)
(9 ,0.234783422)
(10,0.286776802)
(11,0.264259861)
(12,0.211300045)};          
          \end{axis}}

        \end{tikzpicture}
    \end{minipage}
    }
    \subfigure[En-Fr]{
    \begin{minipage}[t]{0.232\linewidth}
    \centering
    \begin{tikzpicture}
        \pgfplotsset{set layers}
         \scriptsize{
         \begin{axis}
        [
       at={(0,0)},
          ymajorgrids,
          xmajorgrids,
          grid style=dashed,
          width=1\textwidth,
          height=1\textwidth,
          legend style={at={(1.8,1.2)}, anchor=south west},
          ylabel={\scriptsize{Localness}},
          ylabel style={yshift=-1.5em},
          yticklabel style={/pgf/number format/fixed zerofill, /pgf/number format/fixed,/pgf/number format/sci precision=2},
          ymin=0.1,ymax=0.73, ytick={0.2,0.3,0.4,0.5,0.6},
          xmin=0,xmax=13,xtick={2,4,6,8,10,12},
          legend columns=4,
          legend style={yshift=-9pt,xshift=-10.66em, legend plot pos=left,cells={anchor=west}}
          ]
          \addplot[red!60,mark=square*,mark size=1.2pt,thick,mark options={fill=white,draw=red,line width=0.5pt}] coordinates
          {
(1 ,0.303570304)
(2 ,0.386140251)
(3 ,0.27733595)
(4 ,0.486639317)
(5 ,0.271179541)
(6 ,0.522426518)
(7 ,0.352821701)
(8 ,0.434835271)
(9 ,0.435675238)
(10,0.253609141)
(11,0.269080571)
(12,0.180478346)}; 

          \addplot[blue!70,mark=diamond*,mark size=1.5pt,thick,mark options={fill=white,draw=blue,line width=0.5pt}] coordinates {
(1 ,0.391281238)
(2 ,0.296069275)
(3 ,0.262322075)
(4 ,0.453776632)
(5 ,0.511211404)
(6 ,0.34226119)
(7 ,0.214852072)
(8 ,0.43937232)
(9 ,0.221152875)
(10,0.273868697)
(11,0.221514042)
(12,0.20272308)}; 
          \end{axis}}

        \end{tikzpicture}
    \end{minipage}
    }

    \captionsetup{skip=2pt}
    \caption{Localness of attention weight with/without task prompt on different language tasks.}
    \label{promptlocalness}
    \vspace{-0.2cm}
\end{figure*}
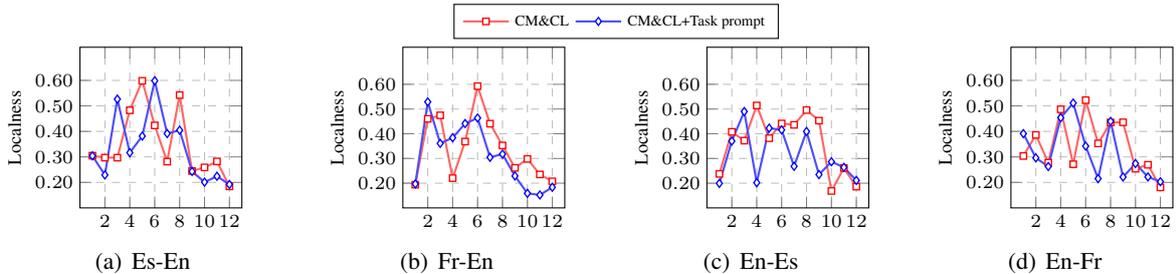
\subsection{How CM and CL Work?}

We have observed that both $\mathcal{L}_{\mathrm{CM}}$ and $\mathcal{L}_{\mathrm{CL}}$ contribute to textless S2ST, then we reveal how these two losses affect the model. We sample 200 audios from the training dataset to compute the \textit{Sparseness} metric and analyze the effect on S2ST models. Sparseness is calculated by determining the proportion of values with absolute values less than 1e-3 in the representations. The representations are extracted from the normalized output of each encoder layer. This metric mainly measures the number of non-activated nodes in the representation, with higher sparseness indicating more unnecessary information. To simplify the expression, we use CM and CL to denote $\mathcal{L}_{\mathrm{CM}}$ and $\mathcal{L}_{\mathrm{CL}}$ separately in the later analysis section. Similarly, CM$^{'}$ and CL$^{'}$ denote models trained based on the recognized text.

Figure \ref{unit_language_influence} illustrates the Sparseness of different layers resulting from different methods. We observe distinct trends for CL and CM. Compared with middle layers, Baseline and CM exhibit higher Sparseness in earlier and top layers. This shows more nodes are not needed for the bottom encoder and target decoder, suggesting its effectiveness in noise filtering but limited cross-lingual understanding. The difference is the slight fluctuation in the 7-th layer for CM. After applying CL, the representation learned other information which differs from the source unit. Conversely, CL does not show an increase in Sparseness in top layers, indicating its better cross-lingual capability. 

We test the model Sparseness after applying the text and show results in Figure \ref{text_influence}. Both tendencies of unit language and text are consistent, suggesting the unit language could play the role of text. This also confirms the effect of CL and CM is distinct and explain why both the two tasks work well.

\subsection{Why CM and CL ``Conflict''?}
Previous work identified a gap between CM and CL in speech-to-text translation \cite{xu-etal-2021-stacked}. We examine if this issue also causes inconsistency here with the \textit{localness} metric. The metric sums self-attention weights within a window, and lower localness indicates global attention focus. We randomly sample 200 audio recordings from the training dataset and extract attention weights with a window size of 10 in each layer.

Figure \ref{localness} shows the results of different models. The localness of the baseline increases initially and reaches its peak at layers 4 to 6 before decreasing. This indicates that S2ST requires cross-modal processing first, followed by cross-lingual processing. CM works in the middle layer and shows higher localness. CL in the upper layers shows higher locality, which is affected by cross-lingual learning. Considering the conflict of using the two together, our conjecture is that the guidance of CM in the middle layer affects the learning of CL at the top level, resulting in the two tasks being unable to help each other. Therefore, we designed task prompts to alleviate the learning conflict.

\subsection{How to Make CM and CL Harmonious?}

We show the change in localness after applying the task prompt in Figure \ref{promptlocalness}. Although the overall tendency of the two methods are completely different, it can be observed that the consistent changes come from higher layers. When not using task prompt, there will be a significant concussion on the around 8th layer, which we think is the disturbance caused by CM to CL. After applying prompt modeling, CL is easier to learn the semantic information, thus localness is lower at the top level and more inclined to learn global information. Thus designing a strategy to avoid the conflict is necessary.

\begin{table}[t]
    \centering
    \small
    \setlength{\tabcolsep}{1.5mm}{
    \begin{tabular}{lcccccc}
    \toprule
     Models &\multicolumn{3}{c}{En-Es}&\multicolumn{3}{c}{En-Fr} \\ \cmidrule(r){2-4} \cmidrule(r){5-7}
     &$r$=0&$r$=2&$r$=4&$r$=0&$r$=2&$r$=4\\
     
    \midrule
    Baseline &23.0 &23.0&23.0 &18.8 &18.8&18.8 \\
    \ \ + CM & 21.7&23.5 &22.3 &17.9&19.5& 19.0\\ 
    \ \ + CM\&CL &23.1&\textbf{23.8}&23.8&19.6&\textbf{20.4}&19.8 \\ 
    \bottomrule
    \end{tabular}}
    \caption{The performance of different T-Enc layers in applying the CL training.}
    \label{r_hyper_param}
    \vspace{-0.3cm}
\end{table}

\subsection{Effect of Hyper Parameters}
There are two hyperparameters here, $r$ and $K$. We first test adding the $\mathcal{L}_{\mathrm{CM}}$ at three different layers, and the results are shown in Table \ref{r_hyper_param}. If the CM training is applied in the same layer as source unit training, the two tasks conflict and hurt the performance significantly. This proves the effect of unit language differs greatly from that of the unit. Additionally, if the CM training is applied near the CL training, the performance degrades. This also confirms the previous conclusion that there is a conflict between CL and CM training.

Table \ref{k_hyper_param} shows the results for different values of $K$. $K$ represents the maximum size used to build the unit language. We find that $K$ is highly related to the language pairs, as each language has its unique pronunciation units. Furthermore, there is a threshold for each language pair, meaning that increasing the size of $K$ beyond this threshold does not improve performance. This confirms that unit language is highly related to the language features.

\subsection{Comparison with BPE Method}
Previous work \cite{shen2024acoustic} used the BPE method to generate a pseudo language. We reproduced this method based on our multi-task training and normalized units to compared with the proposed unit langauge. We found that the unit language significantly outperforms the BPE method, as shown in Table \ref{bpe}. This is because the BPE method does not consider phrase information, while our method applies $n$-gram modeling, which makes the unit language more accurate.

\section{Related Work}
\citet{jia19_interspeech} successfully built S2ST with auxiliary text tasks. Another approach to constructing direct S2ST is by combining advanced end-to-end speech translation methods with unit-based text-to-speech methods \cite{inaguma-etal-2023-unity, barrault2023seamlessm4t}. Due to the scarcity of training data, some studies utilize unsupervised methods or data augmentation to enhance performance \cite{jia22b_interspeech, dong22b_interspeech, popuri22_interspeech}. A challenge in textless S2ST is extracting acoustic and semantic features from noisy speech sequences, leading many studies to employ the VQ-VAE method to aid alignment learning between different language speeches \cite{s2svae, zhang2020uwspeech}. Conversely, \citet{lee-etal-2022-direct, lee-etal-2022-textless, chen-etal-2023-speech} regard unit tokens as language text. Our analysis aims to further explore the next steps in unit-based S2ST. Some studies focus on the voice, style, and speed of speech synthesis \cite{Jia2021Translatotron2H, song23_interspeech, huang2022transpeech, Fang2023DASpeechDA}, while our goal is to enhance the modeling ability of S2ST. Related works generate pseudo language based on the byte-pair encoding method \cite{wu2023wav2seq, shen2024acoustic}. Their work focuses on speech-to-text or text-to-speech tasks, while our work aims to improve the more complex textless speech-to-speech task. Furthermore, their method does not consider in-context information when building the language.

\begin{table}[t]
    \centering
    \small
    \setlength{\tabcolsep}{1.5mm}{
    \begin{tabular}{llll|lll}
    \toprule
     Models &\multicolumn{3}{c}{En-Es}&\multicolumn{3}{c}{En-Fr} \\ \cmidrule(r){2-4} \cmidrule(r){5-7}
     &$K$=2&$K$=3&$K$=4&$K$=2&$K$=3&$K$=4\\
    \midrule
    Baseline & 23.0 &23.0&23.0 &18.8 &18.8&18.8 \\
    \ \ + CM &23.3 &23.5 & 23.7 &19.5&19.5& 19.5\\ 
    \ \ + CL &23.1&23.9 & \textbf{24.0} &\textbf{19.9}&\textbf{19.9}& 19.8\\ 
    \bottomrule
    \end{tabular}}
    \caption{The performance of different values of $K$ in generating the unit language.}
    \label{k_hyper_param}
\end{table}

\begin{table}[t]
    \centering
    \small
    \setlength{\tabcolsep}{0.8mm}{
    \begin{tabular}{llllll}
    \toprule
    Models &\multicolumn{1}{c}{Es-En}&\multicolumn{1}{c}{Fr-En}&\multicolumn{1}{c}{En-Es}&\multicolumn{1}{c}{En-Fr}&\multicolumn{1}{c}{Avg.} \\
    \midrule
    Baseline&19.1&20.3&23.0&18.8&20.3 \\
    \ \ + BPE &19.6 & 20.3&23.0 &19.8&20.7 ($+$0.4)\\
    \ \ + Unit language&\textbf{19.7} &\textbf{21.0} &\textbf{23.8} &\textbf{20.4} &\textbf{21.2} ($+$0.9)\\
    \bottomrule
    \end{tabular}}
    \caption{Comparison of BPE method \cite{shen2024acoustic} and unit language.}
    \label{bpe}
    \vspace{-0.2cm}
\end{table}
\section{Conclusion}
Textless S2ST has attracted significant attention from researchers, yet it encounters cross-modal and cross-lingual challenges that impede performance improvement. We introduce unit language to boost either cross-modal or cross-lingual modeling of S2ST. Our analyses demonstrate that CM enhances speech modeling, while CL enhances semantic understanding. We further propose task prompt learning to mitigate conflicts between CM and CL training. Our method achieves comparable performance of textless S2ST to text-based models.

\section*{Acknowledgement}
This work was supported in part by the National Science Foundation of China (Nos. 62276056 and U24A20334), the Natural Science Foundation of Liaoning Province of China (2022-KF-26-01), the Fundamental Research Funds for the Central Universities (Nos. N2216016 and N2316002), the Yunnan Fundamental Research Projects (No. 202401BC070021), and the Program of Introducing Talents of Discipline to Universities, Plan 111 (No.B16009). The authors would like to thank anonymous reviewers for their comments.

\section*{Limitations}
Our work discusses the modeling challenge we are facing in textless speech-to-speech translation and indicates the conflict between CM and CL. However, the investigation of the proper method to solve the conflict is limited. We introduce the task prompt that could mitigate this conflict with source and target text, but the conflict is not completely cleared. Additionally, while we focus on the translation performance, our work lacks human evaluation which could test the tone and fluency.

\bibliography{acl_latex}

\newpage
\appendix

\section*{Appendix}

\section{Data Details}

We have carried out experiments on four languages of Voxpupil, and the specific data of each language is shown in the Table \ref{DataSize}. 

\begin{table}[h]
    \centering
    \setlength{\tabcolsep}{2.0mm}{
    \begin{tabular}{lcc}
    \toprule
    Language&Hours(h)& Sentence(k)\\
    \midrule
    Es-En&530&158 \\
    Fr-En&521&155 \\
    En-Es&413&125 \\
    En-Fr&444&135 \\
    \bottomrule
    \end{tabular}}
    \caption{Training data size of the VoxPopuli 4 languages.}
    \label{DataSize}
\end{table}

\section{Hyper Parameters of Language Modeling}
We primarily apply a 2-gram model here though our method can easily be extended to higher n-gram models. This is because that the computation process becomes very complex and time-consuming with higher n-grams. Table \ref{time} shows the time consumption with the increase of $K$ for one language. Higher values of $K$ require significantly longer computation times to estimate the probabilities accurately. We found that $K=3$ is sufficient for most languages according to the previous analysis. In summary, the current selection of $n$-gram and $K$ to merge units is based on a balance of efficiency and empirical study. 

\begin{table}[h]
    \centering
    \setlength{\tabcolsep}{2.0mm}{
    \begin{tabular}{lc}
    \toprule
    $K$&Time\\
    \midrule
    2& $\sim$2 hours \\ 
    3& $\sim$12 hours \\
    4 (with pruning strategy)& 	$\sim$2 days \\
    \bottomrule
    \end{tabular}}
    \caption{Time consuming of producing the unit language. The size of one corpus is about 160 million units.}
    \label{time}
\end{table}

\section{Necessity of Additional Decoders}
Another alternative way to use the unit language is with CTC \cite{CTC} to predict the objective, which can avoid the need for additional decoders. We compare the two methods, namely CTC and cross-entropy (CE) based on additional decoders, in Table \ref{ctc_loss}. We found that the additional decoders stabilize the whole modeling process, while CTC fails to take advantage of the unit language.

\begin{table}[h]
    \centering
    \setlength{\tabcolsep}{0.6mm}{
    \begin{tabular}{lll}
    \toprule
    \multirow{2}*{}&\multirow{2}*{CTC loss} & Decoder  \\
    & & $w/$ CE loss\\
    \midrule
Baseline &23.0 &23.0\\
\ \ +Src unit langauge &22.8(-0.2) &23.9(+0.9)\\
\ \ +Tgt unit langauge &17.3(-5.7) &23.9(+0.9)\\
    \bottomrule
    \end{tabular}}
    \caption{Using CTC loss or CE loss to leverage the unit language on En-Es task.}
    \label{ctc_loss}
\end{table}

\section{Task Prompt Based on CM$^{'}$ and CL$^{'}$}
We test the task prompt training based on $\mathcal{L}_{\mathrm{CM}^{'}}$ and $\mathcal{L}_{\mathrm{CL}^{'}}$. Results shown in Table \ref{text_prompt} show that the task prompt still works well and the conflict between CL and CM is not caused by the unit language.

\section{Sparseness Results on Spanish}
Due to the page limit, we show the effect of the model in Spanish after using CL and CM respectively in the Figure \ref{ul_influence_es}. It can be seen that the conclusion is consistent with the French phenomenon described in the main content. We also exhibit Sparseness results with text in Figure \ref{text_influence_es}. The phenomena are the same as those in the previous experiments.

\section{Localness Results Based on CM$^{'}$ and CL$^{'}$}
The localness results of models which apply the CM$^{'}$ and CL$^{'}$ are shown in Figure \ref{localness_text}. The trend shown in the figure is consistent with the use of CL and CM methods described in the text, demonstrating that both methods achieve the goals of cross-lingual learning and cross-modal learning.

\section{Effect of Norm Unit}
The units used in this work have been normalized \citet{lee-etal-2022-textless} to reduce noise and other unimportant information. This baseline method in our work achieves about a 3 BLEU improvement, highlighting the importance of unit normalization. If the units are not normalized, unit language does not perform well as shown in Table \ref{Norm}.

\begin{figure*}[p]
    \centering
    \subfigure[Es-En]{
    \begin{minipage}[t]{0.485\linewidth}
    \centering
    \begin{tikzpicture}
  
        \pgfplotsset{set layers}
         \scriptsize{
         \begin{axis}
        [
       at={(0,0)},
          ymajorgrids,
          xmajorgrids,
          grid style=dashed,
          width=1.\textwidth,
          height=.5\textwidth,
          legend style={at={(0.3,0.78)}, anchor=south west},
          ylabel={\scriptsize{Sparseness}},
          ylabel style={yshift=-1.5em},
          yticklabel style={/pgf/number format/fixed zerofill, /pgf/number format/fixed,/pgf/number format/sci precision=2},
          ymin=0.002,ymax=0.0055, ytick={0.0025,0.003,0.0035,0.004,0.0045,0.005},
          xmin=0,xmax=13,xtick={1,2,3,4,5,6,7,8,9,10,11,12},
          legend columns=4,
          legend style={yshift=-1pt,xshift=-4em, legend plot pos=left,cells={anchor=west}}
          ]
          \addplot[blue!60,mark=pentagon*,mark size=1.5pt,thick,mark options={fill=white,draw=blue,line width=0.5pt}] coordinates {
(1 ,0.00387263238732464)
(2 ,0.00324707921890027)
(3 ,0.00277944462408383)
(4 ,0.00267609799398069)
(5 ,0.00250690877566079)
(6 ,0.00259921146942261)
(7 ,0.00250995314520943)
(8 ,0.00235951594783272)
(9 ,0.0024571300948496)
(10 ,0.00247879693770105)
(11 ,0.00255859828416741)
(12 ,0.00356670563459688)}; 

          \addlegendentry{\scalebox{.8}{Baseline}}
          \addplot[teal!70,mark=diamond*,mark size=1.5pt,thick,mark options={fill=white,draw=teal,line width=0.5pt}] coordinates {
(1 ,0.00477995167354823)
(2 ,0.00400710283802606)
(3 ,0.00393021631346798)
(4 ,0.00320640125982489)
(5 ,0.00301771512162968)
(6 ,0.00318237017253672)
(7 ,0.00319548687112393)
(8 ,0.00247685372309554)
(9 ,0.00236702971097403)
(10 ,0.00272658918681391)
(11 ,0.00314680934525586)
(12 ,0.00352586574097104)};

          \addlegendentry{\scalebox{.8}{CM}}
          \addplot[orange!80,mark=triangle*,,mark size=1.5pt,thick,mark options={fill=white,draw=orange,line width=0.5pt}] coordinates {
(1 ,0.00373006520910025)
(2 ,0.00319244250157529)
(3 ,0.00296168576717076)
(4 ,0.0027560936619076)
(5 ,0.00256054149877292)
(6 ,0.00263937123793652)
(7 ,0.00260186719605014)
(8 ,0.00239060738152091)
(9 ,0.00234354920115743)
(10 ,0.00228091291703976)
(11 ,0.00229529270512055)
(12 ,0.00225934323491858)};

          \addlegendentry{\scalebox{.8}{CL}}
          \addplot[red!60,mark=square*,mark size=1.2pt,thick,mark options={fill=white,draw=red,line width=0.5pt}] coordinates {(1 ,0.005123956) (2 ,0.004205119) (3 ,0.003827618) (4 ,0.003334735) (5 ,0.003151047) (6 ,0.002822168) (7 ,0.00279541) (8 ,0.002452212) (9 ,0.002319158) (10,0.002312284) (11,0.002539864) (12,0.002563403)};
          \addlegendentry{\scalebox{.8}{CM\&CL}}
  
          \end{axis}}

        \end{tikzpicture}
    \end{minipage}
    }
    \subfigure[En-Es]{
    \begin{minipage}[t]{0.485\linewidth}
    \centering
\begin{tikzpicture}
        \pgfplotsset{set layers}
         \scriptsize{
         \begin{axis}
        [
       at={(0,0)},
          ymajorgrids,
          xmajorgrids,
          grid style=dashed,
          width=1\textwidth,
          height=.5\textwidth,
          legend style={at={(0.3,0.78)}, anchor=south west},
          ylabel={\scriptsize{Sparseness}},
          ylabel style={yshift=-1.5em},
          yticklabel style={/pgf/number format/fixed zerofill, /pgf/number format/fixed,/pgf/number format/sci precision=2},
          ymin=0.002,ymax=0.0055, ytick={0.0025,0.003,0.0035,0.004,0.0045,0.005},
          xmin=0,xmax=13,xtick={1,2,3,4,5,6,7,8,9,10,11,12},
          legend columns=4,
          legend style={yshift=-1pt,xshift=-4em, legend plot pos=left,cells={anchor=west}}
          ]
          \addplot[blue!60,mark=pentagon*,mark size=1.5pt,thick,mark options={fill=white,draw=blue,line width=0.5pt}] coordinates {
(1 ,0.00378617273207203)
(2 ,0.00309149505553934)
(3 ,0.00278032023108486)
(4 ,0.00258409231376723)
(5 ,0.00243507712843579)
(6 ,0.00262459869177811)
(7 ,0.00255984884414647)
(8 ,0.00245221186284272)
(9 ,0.00234933639286143)
(10,0.00247883608812471)
(11,0.00259670025924081)
(12,0.00355537356001167)}; 

          \addlegendentry{\scalebox{.8}{Baseline}}
          \addplot[teal!70,mark=diamond*,mark size=1.5pt,thick,mark options={fill=white,draw=teal,line width=0.5pt}] coordinates {
(1 ,0.00415542458195271)
(2 ,0.00373252190026954)
(3 ,0.00316375601962332)
(4 ,0.00295844099266915)
(5 ,0.00268602889419197)
(6 ,0.00268901322171098)
(7 ,0.00290224174635603)
(8 ,0.00239903047581849)
(9 ,0.00234192587171871)
(10,0.00241951838721307)
(11,0.00275845751712534)
(12,0.00361066744853812)}; 
        
          \addlegendentry{\scalebox{.8}{CM}}
          \addplot[orange!80,mark=triangle*,,mark size=1.5pt,thick,mark options={fill=white,draw=orange,line width=0.5pt}] coordinates {
(1 ,0.00417809205838927)
(2 ,0.00346822449010249)
(3 ,0.00332819581909797)
(4 ,0.00307607044139612)
(5 ,0.00294486062586914)
(6 ,0.00283822959766168)
(7 ,0.00284731670729823)
(8 ,0.00272576404149569)
(9 ,0.00254982284495338)
(10,0.00241636640084467)
(11,0.00237009255841502)
(12,0.00251887302135732)};

          \addlegendentry{\scalebox{.8}{CL}}
          \addplot[red!60,mark=square*,mark size=1.2pt,thick,mark options={fill=white,draw=red,line width=0.5pt}] coordinates { (1 ,0.004748633531489404) (2 ,0.004284237627682154) (3 ,0.0037735608293536296) (4 ,0.0033007119420124034) (5 ,0.003016613966686565) (6 ,0.0029001506379962194) (7 ,0.0029807940441249625) (8 ,0.0025828884667363117) (9 ,0.0024569357733890494) (10,0.0023217528106656052) (11,0.0025909851942592776) (12,0.0024418110863761483)};
          \addlegendentry{\scalebox{.8}{CM\&CL}}
  
          \end{axis}}

        \end{tikzpicture}
    \end{minipage}
    }
    \captionsetup{skip=2pt}
    \caption{The influence of CM and CL on Es-En (left) and En-Es (right) tasks.}
    \label{ul_influence_es}
\end{figure*}


\begin{figure*}[p]
    \centering
    \subfigure[Es-En]{
    \begin{minipage}[t]{0.485\linewidth}
    \centering
    \begin{tikzpicture}
  
        \pgfplotsset{set layers}
         \scriptsize{
         \begin{axis}
        [
       at={(0,0)},
          ymajorgrids,
          xmajorgrids,
          grid style=dashed,
          width=1.\textwidth,
          height=.5\textwidth,
          legend style={at={(0.3,0.78)}, anchor=south west},
          ylabel={\scriptsize{Sparseness}},
          ylabel style={yshift=-1.5em},
          yticklabel style={/pgf/number format/fixed zerofill, /pgf/number format/fixed,/pgf/number format/sci precision=2},
          ymin=0.002,ymax=0.0055, ytick={0.0025,0.003,0.0035,0.004,0.0045,0.005},
          xmin=0,xmax=13,xtick={1,2,3,4,5,6,7,8,9,10,11,12},
          legend columns=4,
          legend style={yshift=-1pt,xshift=-4em, legend plot pos=left,cells={anchor=west}}
          ]
          \addplot[blue!60,mark=pentagon*,mark size=1.5pt,thick,mark options={fill=white,draw=blue,line width=0.5pt}] coordinates {
(1 ,0.003786173)
(2 ,0.003091495)
(3 ,0.002780320)
(4 ,0.002584025)
(5 ,0.002435144)
(6 ,0.002624699)
(7 ,0.002559849)
(8 ,0.002452145)
(9 ,0.002349135)
(10,0.002478803)
(11,0.002596633)
(12,0.003555374)}; 

          \addlegendentry{\scalebox{.8}{Baseline}}
          \addplot[teal!70,mark=diamond*,mark size=1.5pt,thick,mark options={fill=white,draw=teal,line width=0.5pt}] coordinates {
(1 ,0.004024919)
(2 ,0.003512486)
(3 ,0.003189274)
(4 ,0.002820156)
(5 ,0.002577453)
(6 ,0.00234595)
(7 ,0.002400472)
(8 ,0.002233786)
(9 ,0.002146234)
(10,0.002244617)
(11,0.002433233)
(12,0.00338919)}; 
          
          \addlegendentry{\scalebox{.8}{CM$^{'}$}}
          \addplot[orange!80,mark=triangle*,,mark size=1.5pt,thick,mark options={fill=white,draw=orange,line width=0.5pt}] coordinates {
(1 ,0.004785621)
(2 ,0.004220677)
(3 ,0.004048123)
(4 ,0.003639404)
(5 ,0.0032669)
(6 ,0.003211002)
(7 ,0.003008772)
(8 ,0.002746017)
(9 ,0.002590396)
(10,0.002400707)
(11,0.002627147)
(12,0.002624163)};

          \addlegendentry{\scalebox{.8}{CL$^{'}$}}
          \addplot[red!60,mark=square*,mark size=1.2pt,thick,mark options={fill=white,draw=red,line width=0.5pt}] coordinates {(1 ,0.005123956) (2 ,0.004205119) (3 ,0.003827618) (4 ,0.003334735) (5 ,0.003151047) (6 ,0.002822168) (7 ,0.00279541) (8 ,0.002452212) (9 ,0.002319158) (10,0.002312284) (11,0.002539864) (12,0.002563403)};
          \addlegendentry{\scalebox{.8}{CM$^{'}$\&CL$^{'}$}}
  
          \end{axis}}

        \end{tikzpicture}
    \end{minipage}
    }
    \subfigure[En-Es]{
    \begin{minipage}[t]{0.485\linewidth}
    \centering
\begin{tikzpicture}
        \pgfplotsset{set layers}
         \scriptsize{
         \begin{axis}
        [
       at={(0,0)},
          ymajorgrids,
          xmajorgrids,
          grid style=dashed,
          width=1\textwidth,
          height=.5\textwidth,
          legend style={at={(0.3,0.78)}, anchor=south west},
          ylabel={\scriptsize{Sparseness}},
          ylabel style={yshift=-1.5em},
          yticklabel style={/pgf/number format/fixed zerofill, /pgf/number format/fixed,/pgf/number format/sci precision=2},
          ymin=0.002,ymax=0.0055, ytick={0.0025,0.003,0.0035,0.004,0.0045,0.005},
          xmin=0,xmax=13,xtick={1,2,3,4,5,6,7,8,9,10,11,12},
          legend columns=4,
          legend style={yshift=-1pt,xshift=-4em, legend plot pos=left,cells={anchor=west}}
          ]
          \addplot[blue!60,mark=pentagon*,mark size=1.5pt,thick,mark options={fill=white,draw=blue,line width=0.5pt}] coordinates {
(1 ,0.0038769722332769544)
(2 ,0.0032455894210360495)
(3 ,0.002780707713577422)
(4 ,0.0026762923154412498)
(5 ,0.0025078479960534607)
(6 ,0.0026014137793088582)
(7 ,0.002509694049928697)
(8 ,0.0023605523289556593)
(9 ,0.002453697082379863)
(10,0.0024798333188239974)
(11,0.0025575619030444733)
(12,0.0035593538060060357)}; 

          \addlegendentry{\scalebox{.8}{Baseline}}
          \addplot[teal!70,mark=diamond*,mark size=1.5pt,thick,mark options={fill=white,draw=teal,line width=0.5pt}] coordinates {
(1 ,0.004099502692518157)
(2 ,0.0032407313845222697)
(3 ,0.0031562663230026862)
(4 ,0.0028257579055152057)
(5 ,0.0025400405846847745)
(6 ,0.0023150487202765893)
(7 ,0.002370495110353862)
(8 ,0.0022613512233442775)
(9 ,0.002162279665373263)
(10,0.00228518798917189)
(11,0.002477307139836832)
(12,0.003250026427718635)}; 
        
          \addlegendentry{\scalebox{.8}{CM$^{'}$}}
          \addplot[orange!80,mark=triangle*,,mark size=1.5pt,thick,mark options={fill=white,draw=orange,line width=0.5pt}] coordinates {
(1 ,0.004810945946506152)
(2 ,0.004507091956024276)
(3 ,0.003951041096657049)
(4 ,0.003657971947235764)
(5 ,0.0034667920169634864)
(6 ,0.003160800490415547)
(7 ,0.003108171761516267)
(8 ,0.0028548413507777)
(9 ,0.0027094888982854112)
(10,0.0025516674854077536)
(11,0.0026765190238118926)
(12,0.0025315876011507976)};

          \addlegendentry{\scalebox{.8}{CL$^{'}$}}
          \addplot[red!60,mark=square*,mark size=1.2pt,thick,mark options={fill=white,draw=red,line width=0.5pt}] coordinates { (1 ,0.004748633531489404) (2 ,0.004284237627682154) (3 ,0.0037735608293536296) (4 ,0.0033007119420124034) (5 ,0.003016613966686565) (6 ,0.0029001506379962194) (7 ,0.0029807940441249625) (8 ,0.0025828884667363117) (9 ,0.0024569357733890494) (10,0.0023217528106656052) (11,0.0025909851942592776) (12,0.0024418110863761483)};
          \addlegendentry{\scalebox{.8}{CM$^{'}$\&CL$^{'}$}}
  
          \end{axis}}

        \end{tikzpicture}
    \end{minipage}
    }
    \captionsetup{skip=2pt}
    \caption{The influence of CM$^{'}$ and CL$^{'}$ on Es-En (left) and En-Es (right) tasks.}
    \label{text_influence_es}
\end{figure*}

\begin{figure*}[p]
    \centering
    \subfigure[Es-En]{
    \begin{minipage}[t]{0.232\linewidth}
    \centering
    \begin{tikzpicture}
        \pgfplotsset{set layers}
         \scriptsize{
         \begin{axis}
        [
       at={(0,0)},
          ymajorgrids,
          xmajorgrids,
          grid style=dashed,
          width=1\textwidth,
          height=1\textwidth,
          legend style={at={(1.8,1.2)}, anchor=south west},
          ylabel={\scriptsize{Localness}},
          ylabel style={yshift=-1.5em},
          yticklabel style={/pgf/number format/fixed zerofill, /pgf/number format/fixed,/pgf/number format/sci precision=2},
          ymin=0.07,ymax=0.63, ytick={0.2,0.3,0.4,0.5},
          xmin=1,xmax=13,xtick={2,4,6,8,10,12},
          legend columns=4,
          legend style={yshift=-9pt,xshift=1.48em, legend plot pos=left,cells={anchor=west}}
          ]
          \addplot[blue!60,mark=pentagon*,mark size=1.5pt,thick,mark options={fill=white,draw=blue,line width=0.5pt}] coordinates {
(2 ,0.364435303)
(4 ,0.429218289)
(6 ,0.489436356)
(8 ,0.284973698)
(10,0.134021130)
(12,0.128955043)}; 

          \addlegendentry{\scalebox{.8}{Baseline}}
          \addplot[teal!70,mark=diamond*,mark size=1.5pt,thick,mark options={fill=white,draw=teal,line width=0.5pt}] coordinates {
(2 ,0.420004473)
(4 ,0.410841695)
(6 ,0.576016994)
(8 ,0.206919965)
(10,0.173838859)
(12,0.140244854)}; 

          \addlegendentry{\scalebox{.8}{CM$^{'}$}}
          \addplot[orange!80,mark=triangle*,,mark size=1.5pt,thick,mark options={fill=white,draw=orange,line width=0.5pt}] coordinates {
(2 ,0.231354128)
(4 ,0.380202018)
(6 ,0.46398827)
(8 ,0.314994462)
(10,0.396346255)
(12,0.221859924)};

          \addlegendentry{\scalebox{.8}{CL$^{'}$}}
          \addplot[red!60,mark=square*,mark size=1.2pt,thick,mark options={fill=white,draw=red,line width=0.5pt}] coordinates {
(2 ,0.355424062)
(4 ,0.555958711)
(6 ,0.390890856)
(8 ,0.168025533)
(10,0.301796989)
(12,0.230426466)};

          \addlegendentry{\scalebox{.8}{CM$^{'}$\&CL$^{'}$}}
          \end{axis}}

        \end{tikzpicture}
    \end{minipage}
    }
    \subfigure[Fr-En]{
    \begin{minipage}[t]{0.232\linewidth}
    \centering
    \begin{tikzpicture}
        \pgfplotsset{set layers}
         \scriptsize{
         \begin{axis}
        [
       at={(0,0)},
          ymajorgrids,
          xmajorgrids,
          grid style=dashed,
          width=1\textwidth,
          height=1\textwidth,
          legend style={at={(0.01,0.91)}, anchor=south west},
          ylabel={\scriptsize{Localness}},
          ylabel style={yshift=-1.5em},
          yticklabel style={/pgf/number format/fixed zerofill, /pgf/number format/fixed,/pgf/number format/sci precision=2},
          ymin=0.05,ymax=0.75, ytick={0.2,0.3,0.4,0.5,0.6},
          xmin=1,xmax=13,xtick={2,4,6,8,10,12},
          legend columns=4,
          legend style={yshift=-9pt,xshift=0.5em, legend plot pos=left,cells={anchor=west}}
          ]
          \addplot[blue!60,mark=pentagon*,mark size=1.5pt,thick,mark options={fill=white,draw=blue,line width=0.5pt}] coordinates {
(2 ,0.368665302)
(4 ,0.582022339)
(6 ,0.446468657)
(8 ,0.293454195)
(10,0.117980306)
(12,0.112620136)}; 
          \addplot[teal!70,mark=diamond*,mark size=1.5pt,thick,mark options={fill=white,draw=teal,line width=0.5pt}] coordinates {
(2 ,0.450516105)
(4 ,0.529115924)
(6 ,0.464232664)
(8 ,0.206125741)
(10,0.180656948)
(12,0.160928385)}; 
   
          \addplot[orange!80,mark=triangle*,,mark size=1.5pt,thick,mark options={fill=white,draw=orange,line width=0.5pt}] coordinates {
(2 ,0.295568275)
(4 ,0.334514883)
(6 ,0.393968864)
(8 ,0.259991345)
(10,0.335683186)
(12,0.19422569)};

          \addplot[red!60,mark=square*,mark size=1.2pt,thick,mark options={fill=white,draw=red,line width=0.5pt}] coordinates {
(2 ,0.356332992)
(4 ,0.404617354)
(6 ,0.531856783)
(8 ,0.198556957)
(10,0.15968295)
(12,0.176869414)};
          \end{axis}}

        \end{tikzpicture}
    \end{minipage}
    }
    \subfigure[En-Es]{
    \begin{minipage}[t]{0.232\linewidth}
    \centering
    \begin{tikzpicture}
        \pgfplotsset{set layers}
         \scriptsize{
         \begin{axis}
        [
       at={(0,0)},
          ymajorgrids,
          xmajorgrids,
          grid style=dashed,
          width=1\textwidth,
          height=1\textwidth,
          legend style={at={(0.01,0.91)}, anchor=south west},
          ylabel={\scriptsize{Localness}},
          ylabel style={yshift=-1.5em},
          yticklabel style={/pgf/number format/fixed zerofill, /pgf/number format/fixed,/pgf/number format/sci precision=2},
          ymin=0.15,ymax=0.65, ytick={0.2,0.3,0.4,0.5,0.6},
          xmin=1,xmax=13,xtick={2,4,6,8,10,12},
          legend columns=4,
          legend style={yshift=-9pt,xshift=0.5em, legend plot pos=left,cells={anchor=west}}
          ]
          \addplot[blue!60,mark=pentagon*,mark size=1.5pt,thick,mark options={fill=white,draw=blue,line width=0.5pt}] coordinates {
(2 ,0.311947189)
(4 ,0.395196321)
(6 ,0.518754282)
(8 ,0.294968784)
(10,0.190825579)
(12,0.161509152)}; 
          \addplot[teal!70,mark=diamond*,mark size=1.5pt,thick,mark options={fill=white,draw=teal,line width=0.5pt}] coordinates {
(2 ,0.288136101)
(4 ,0.461454838)
(6 ,0.394108802)
(8 ,0.23618438)
(10,0.220314206)
(12,0.178704616)}; 

          \addplot[orange!80,mark=triangle*,,mark size=1.5pt,thick,mark options={fill=white,draw=orange,line width=0.5pt}] coordinates {
(2 ,0.260417448)
(4 ,0.399443958)
(6 ,0.417838195)
(8 ,0.266543445)
(10,0.25930678)
(12,0.204373766)};

          \addplot[red!60,mark=square*,mark size=1.2pt,thick,mark options={fill=white,draw=red,line width=0.5pt}] coordinates {
(2 ,0.434510166)
(4 ,0.351800261)
(6 ,0.507805864)
(8 ,0.204880418)
(10,0.247994854)
(12,0.177037232)};
          \end{axis}}

        \end{tikzpicture}
    \end{minipage}
    }
    \subfigure[En-Fr]{
    \begin{minipage}[t]{0.232\linewidth}
    \centering
    \begin{tikzpicture}
        \pgfplotsset{set layers}
         \scriptsize{
         \begin{axis}
        [
       at={(0,0)},
          ymajorgrids,
          xmajorgrids,
          grid style=dashed,
          width=1\textwidth,
          height=1\textwidth,
          legend style={at={(0.01,0.91)}, anchor=south west},
          ylabel={\scriptsize{Localness}},
          ylabel style={yshift=-1.5em},
          yticklabel style={/pgf/number format/fixed zerofill, /pgf/number format/fixed,/pgf/number format/sci precision=2},
          ymin=0.05,ymax=0.65, ytick={0.1,0.2,0.3,0.4,0.5,0.6},
          xmin=1,xmax=13,xtick={2,4,6,8,10,12},
          legend columns=4,
          legend style={yshift=-9pt,xshift=0.5em, legend plot pos=left,cells={anchor=west}}
          ]
          \addplot[blue!60,mark=pentagon*,mark size=1.5pt,thick,mark options={fill=white,draw=blue,line width=0.5pt}] coordinates {
(2 ,0.33450858)
(4 ,0.463031033)
(6 ,0.566385098)
(8 ,0.283448474)
(10,0.160751289)
(12,0.169919766)}; 

          \addplot[teal!70,mark=diamond*,mark size=1.5pt,thick,mark options={fill=white,draw=teal,line width=0.5pt}] coordinates {
(2 ,0.247868366)
(4 ,0.409469664)
(6 ,0.389353803)
(8 ,0.118583465)
(10,0.228908233)
(12,0.169632607)}; 
 
          \addplot[orange!80,mark=triangle*,,mark size=1.5pt,thick,mark options={fill=white,draw=orange,line width=0.5pt}] coordinates {
(2 ,0.218551673)
(4 ,0.248082292)
(6 ,0.502139346)
(8 ,0.288593736)
(10,0.360519678)
(12,0.219434852)};

          \addplot[red!60,mark=square*,mark size=1.2pt,thick,mark options={fill=white,draw=red,line width=0.5pt}] coordinates {
(2 ,0.122020686)
(4 ,0.310316107)
(6 ,0.420852441)
(8 ,0.2050648)
(10,0.29575264)
(12,0.184711149)};

          \end{axis}}

        \end{tikzpicture}
    \end{minipage}
    }
    \captionsetup{skip=2pt}
    \caption{Localness of attention weight on different tasks.}
    \label{localness_text}
\end{figure*}


\begin{table*}[t]
    \centering
    \setlength{\tabcolsep}{2.0mm}{
    \begin{tabular}{llllll}
    \toprule
     Models &\multicolumn{1}{c}{Es-En}&\multicolumn{1}{c}{Fr-En}&\multicolumn{1}{c}{En-Es}&\multicolumn{1}{c}{En-Fr}&\multicolumn{1}{c}{Avg.} \\
    \midrule
    Baseline  & 19.1 & 20.3 & 23.0 & 18.8&20.3 \\
    \ \ + $\mathcal{L}_{\mathrm{CM}^{'}}$\&$\mathcal{L}_{\mathrm{CL}^{'}}$&19.8 ($+$0.7)&20.8 ($+$0.5)&24.0 ($+$1.0)&20.8 ($+$2.0)& 21.4 ($+$1.1)\\ 
    \ \ \ \ +Task prompt&\textbf{20.3} ($+$1.2)&\textbf{21.4} ($+$1.1)&\textbf{24.6} ($+$1.6)&\textbf{20.9} ($+$2.1)&\textbf{21.8} ($+$1.5)\\ 
    \bottomrule
    \end{tabular}}
    \caption{Performance on different datasets.} 
    \label{text_prompt}
\end{table*}

\begin{table*}[t]
    \centering
    \setlength{\tabcolsep}{2.0mm}{
    \begin{tabular}{llllll}
    \toprule
    Models &\multicolumn{1}{c}{Es-En}&\multicolumn{1}{c}{Fr-En}&\multicolumn{1}{c}{En-Es}&\multicolumn{1}{c}{En-Fr}&\multicolumn{1}{c}{Avg.} \\
    \midrule
    Baseline&19.1&20.3&23.0&18.8&20.3 \\
    \ \ + $\mathcal{L}_{\mathrm{CM}}$\&$\mathcal{L}_{\mathrm{CL}}$ $w/o$ norm unit &19.3 & 20.3 &22.9 &19.3&20.5 ($+$0.2)\\
    \ \ + $\mathcal{L}_{\mathrm{CM}}$\&$\mathcal{L}_{\mathrm{CL}}$&\textbf{19.7} &\textbf{21.0} &\textbf{23.8 }&\textbf{20.4} &\textbf{21.2} ($+$0.9)\\
    \bottomrule
    \end{tabular}}
    \caption{Comparison of unit language based on norm and un-norm unit.}
    \label{Norm}
\end{table*}


\end{document}